\documentclass[sigconf]{acmart}

\usepackage{url}
\usepackage{caption}
\usepackage{enumerate}
\usepackage{CJK}
\usepackage{enumitem} 
\usepackage{amsmath}

\usepackage{booktabs}

\usepackage{float}
\usepackage{amssymb}
\usepackage{pifont} 
\usepackage[utf8]{inputenc}
\usepackage{tabularx} % For adjustable-width table columns
\usepackage{multirow} % For multi-row cells
\usepackage{graphicx} % Required for including images
\usepackage{xcolor}
\usepackage{colortbl}
\usepackage{subcaption}
\usepackage{cleveref}
\usepackage{algorithm}
\usepackage{algorithmic}
\AtBeginDocument{%
  }

\begin{document}

\title{Friend or Foe? Harnessing Controllable Overfitting for Anomaly Detection}
\author{Long Qian}
\affiliation{
  \institution{Foundation Model Research Center, Institute of Automation, Chinese Academy of Sciences}
  \city{Beijing}
  \country{China}
}
\affiliation{
  \institution{School of Artificial Intelligence, University of Chinese Academy of Sciences}
  \city{Beijing}
  \country{China}
}
\email{qianlong2024@ia.ac.cn}

\author{Bingke Zhu}
\affiliation{
  \institution{Foundation Model Research Center, Institute of Automation, Chinese Academy of Sciences}
  \city{Beijing}
  \country{China}
}
\affiliation{
  \institution{Objecteye Inc.}
  \city{Beijing}
  \country{China}
}
\email{bingke.zhu@nlpr.ia.ac.cn}

\author{Yingying Chen}
\affiliation{
  \institution{Foundation Model Research Center, Institute of Automation, Chinese Academy of Sciences}
  \city{Beijing}
  \country{China}
}
\affiliation{
  \institution{Objecteye Inc.}
  \city{Beijing}
  \country{China}
}
\email{yingying.chen@nlpr.ia.ac.cn}

\author{Ming Tang}
\affiliation{
  \institution{Foundation Model Research Center, Institute of Automation, Chinese Academy of Sciences}
  \city{Beijing}
  \country{China}
}
\affiliation{
  \institution{School of Artificial Intelligence, University of Chinese Academy of Sciences}
  \city{Beijing}
  \country{China}
}
\email{tangm@nlpr.ia.ac.cn}

\author{Jinqiao Wang}
\affiliation{
  \institution{Foundation Model Research Center, Institute of Automation, Chinese Academy of Sciences}
  \city{Beijing}
  \country{China}
}
\affiliation{
  \institution{School of Artificial Intelligence, University of Chinese Academy of Sciences}
  \city{Beijing}
  \country{China}
}
\affiliation{
  \institution{Objecteye Inc.}
  \city{Beijing}
  \country{China}
}
\email{jqwang@nlpr.ia.ac.cn}

\settopmatter{printacmref=false} 
\renewcommand{\shortauthors}{Long Qian et al.}

\begin{abstract}
Overfitting has traditionally been viewed as detrimental to anomaly detection, where excessive generalization often limits models' sensitivity to subtle anomalies. Our work challenges this conventional view by introducing Controllable Overfitting-based Anomaly Detection (COAD), a novel framework that strategically leverages overfitting to enhance anomaly discrimination capabilities. We propose the Aberrance Retention Quotient (ARQ), a novel metric that systematically quantifies the extent of overfitting, enabling the identification of an optimal "golden overfitting interval" $ARQ_{optimal}$ wherein model sensitivity to anomalies is maximized without sacrificing generalization. To comprehensively capture how overfitting affects detection performance, we further propose the Relative Anomaly Distribution Index (RADI), a metric superior to traditional AUROC by explicitly modeling the separation between normal and anomalous score distributions. Theoretically, RADI leverages ARQ to track and evaluate how overfitting impacts anomaly detection, offering an integrated approach to understanding the relationship between overfitting dynamics and model efficacy. We also rigorously validate the statistical efficacy of Gaussian noise as pseudo-anomaly generators, reinforcing the method’s broad applicability. Empirical evaluations demonstrate that our controllable overfitting method achieves State-Of-The-Art(SOTA) performance in both one-class and multi-class anomaly detection tasks, thus redefining overfitting as a powerful strategy rather than a limitation.

\end{abstract}





\maketitle

\section{Introduction}
In the traditional context of machine learning, overfitting has long been perceived as an undesirable phenomenon, just like ~\cite{bishop2006pattern, Goodfellow-et-al-2016, StatisticalLearning, 38136}, traditionally viewed as the result of a model excessively memorizing training data, leading to compromised generalization capabilities. However, we challenge this conventional wisdom by extending the concept beyond the domain of anomaly detection, positioning controllable overfitting as a mechanism for enhancing model sensitivity across a wide range of applications. This leads us to the central question that underpins our work: can overfitting be a friend rather than a foe in the domain of anomaly detection?

\begin{figure}
  \centering
  \begin{subfigure}{0.48\linewidth}
    \includegraphics[width=\linewidth]{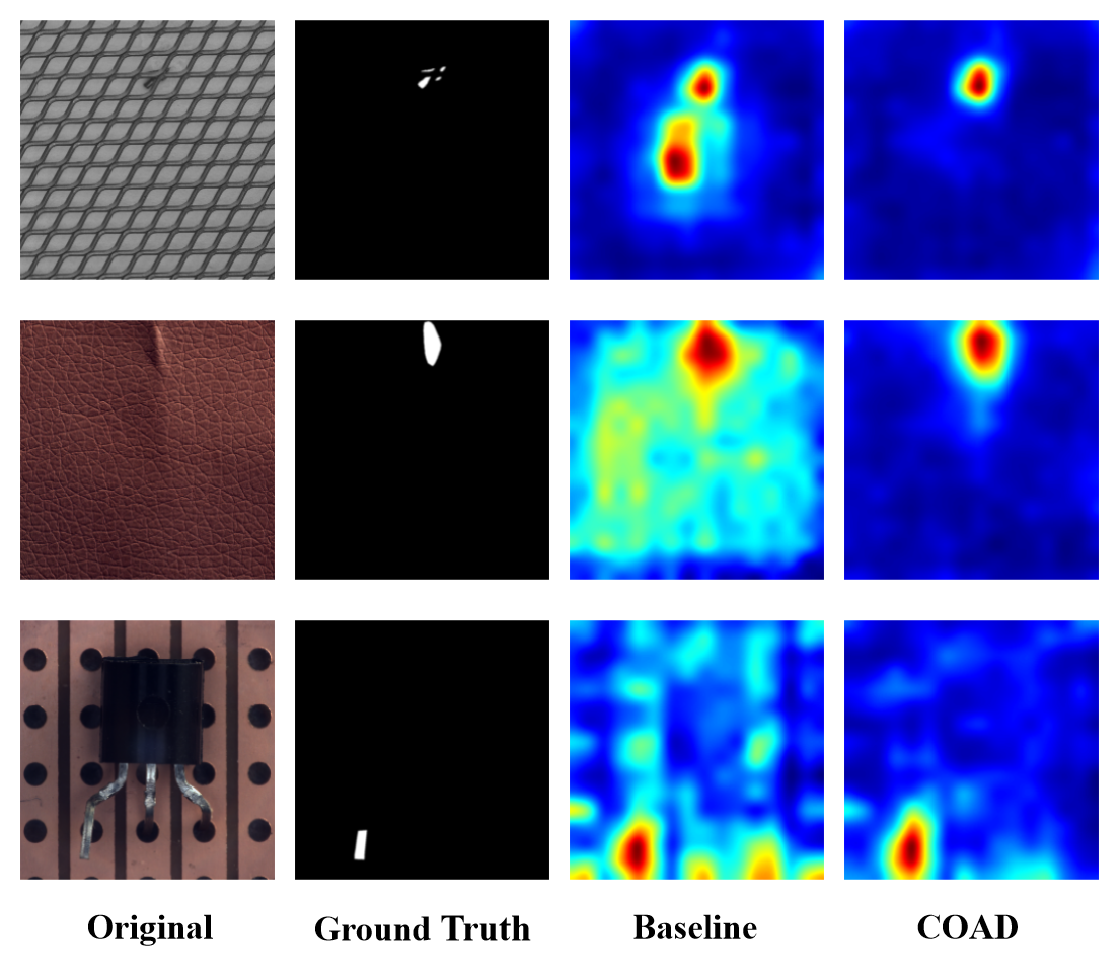}
    \caption{False Positive Cases}
    \label{fig:false_positive}
  \end{subfigure}
  \hfill
  \begin{subfigure}{0.48\linewidth}
    \includegraphics[width=\linewidth]{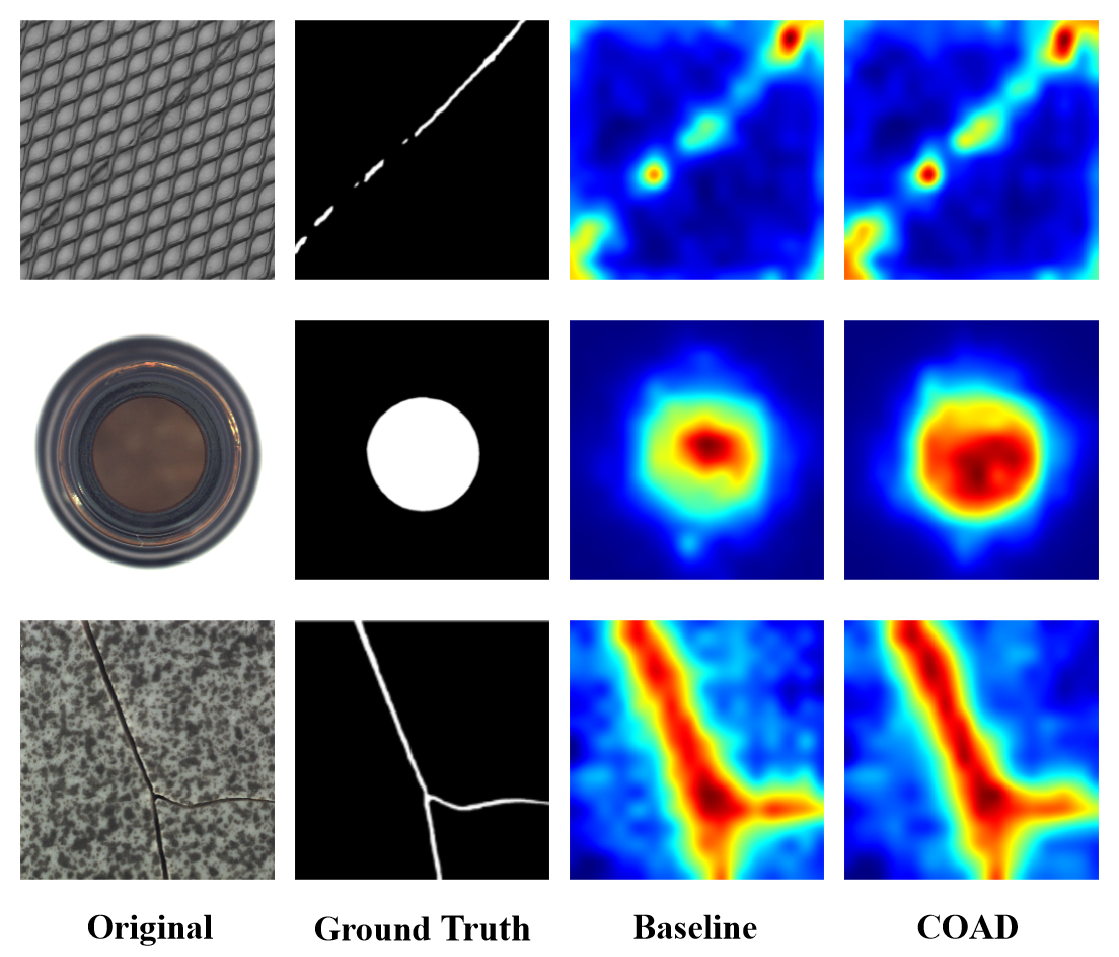}
    \caption{False Negative Cases}
    \label{fig:false_negative}
  \end{subfigure}
  \caption{Each subfigure presents a sequence from left to right, displaying: the original image, the ground truth, the results from other methods, like \textit{RD}~\cite{Deng_2022_CVPR}, \textit{RD++}\cite{Tien_2023_CVPR}, \textit{UniAD}~\cite{you2022unifiedmodelmulticlassanomaly} and \textit{DiAD}~\cite{he2023diaddiffusionbasedframeworkmulticlass}, and our results after overfitting enhancement based on existing frameworks with \textit{COAD}. 
  (a) illustrates a significant reduction in the false positive case after applying our overfitting approach. (b) highlights the reduction in the false negative case, demonstrating enhanced sensitivity towards true anomalies with our method. The consistent improvement across both metrics indicates the effectiveness of our methods in refining detection accuracy.}
  \label{fig:intro}
\end{figure}

The conventional understanding of overfitting necessitates a significant paradigm shift. Historically stigmatized as inherently detrimental to machine learning models, overfitting, as our work demonstrates, can under specific controlled conditions be repurposed as a powerful mechanism for enhancing model performance. Intuitively, by enabling the model to intentionally and moderately memorize fine-grained patterns unique to normal samples, we amplify its sensitivity toward any deviations—precisely the subtle anomalies we aim to detect. This reframes overfitting from a phenomenon to avoid into a strategically leveraged resource.

Anomaly detection is a crucial task that aims to identify patterns that do not conform to expected behavior. Traditionally, anomaly detection relies on the notion of learning the distribution of normal data and distinguishing anomalies by identifying outliers. Various methods have been proposed, including reconstruction-based approaches~\cite{10.1007/978-3-031-19821-2_31,10192551,Zavrtanik_2021_ICCV}, embedding-based models~\cite{xiao2023restrictedgenerativeprojectiononeclass,zhang2024realnetfeatureselectionnetwork,lei2023pyramidflowhighresolutiondefectcontrastive,gudovskiy2021cflowadrealtimeunsupervisedanomaly}, and anomaly synthesis technique~\cite{chen2024unifiedanomalysynthesisstrategy,you2022unifiedmodelmulticlassanomaly,Liu_2023_CVPR,Tien_2023_CVPR}.

We advance the notion of controllable overfitting, suggesting that a calculated and moderate degree of overfitting enables the model to capture intricate and subtle characteristics of the data. This heightened focus allows for a pronounced contrast when confronted with previously unseen deviations—precisely what is required for successful detection tasks, as opposed to a simplistic pursuit of generalization. This approach provides an opportunity to redefine how we evaluate model capabilities in environments devoid of training data that represents the full spectrum of possible scenarios. In this paper, we present \textit{Controllable Overfitting-based Anomaly Detection}, called \textbf{\textit{COAD}}.

Existing anomaly detection metrics such as \textit{AUROC-pixel} predominantly evaluate performance at discrete thresholds, failing to systematically quantify how overfitting impacts model sensitivity across continuous distribution shifts. To fill this critical gap, we introduce the \textit{Aberrance Retention Quotient}(\textit{\textbf{ARQ}}), explicitly measuring the extent of overfitting by quantifying prediction deviations from normal data. Leveraging \textit{ARQ}, we define a "\textit{Golden Overfitting Interval}" (\textbf{\textit{$ARQ_{optimal}$}}), wherein controlled overfitting maximizes anomaly detection sensitivity without compromising model stability. Additionally, we propose the \textit{Relative Anomaly Distribution Index} (\textbf{\textit{RADI}}), an innovative metric derived directly from cumulative distribution functions of normal and anomalous predictions, capturing subtle changes in their overlap and thus providing a more informative assessment than AUROC alone.

Moreover, we substantiate the rationale for employing Gaussian noise in pseudo-anomaly detection modules. Through calculating the Total Variation Distance (TVD) between actual anomaly data and a Gaussian distribution across the entire dataset, we obtain a fair result. This outcome indicates a high level of similarity between true anomalies and a Gaussian distribution, thereby providing a theoretical validation for the common practice of using Gaussian noise as an effective pseudo-anomaly generator in previous works like \textit{GLASS}~\cite{chen2024unifiedanomalysynthesisstrategy} and \textit{SimpleNet}~\cite{Liu_2023_CVPR}. This insight not only strengthens our methodology but also offers a theoretical foundation for broader applications in anomaly detection frameworks.

To summarize, the key contributions of our work are as follows:

\begin{itemize} 
\item We challenge and transcend the conventional understanding of overfitting, positioning it as a controllable and transformative mechanism capable of unlocking model capabilities beyond traditional boundaries. 
\item We introduce the \textit{ARQ} to precisely regulate the overfitting degree and the \textit{RADI} as a complementary metric to \textit{AUROC-pixel}, providing a more adaptable tool for mathematical modeling and facilitating \textit{Dual Control Mechanism}. 
\item Our method achieves SOTA results in both \textit{one-class} and \textit{multi-class} anomaly detection tasks. We also provide a theoretical foundation for utilizing \textit{Gaussian noise} as a preliminarily pseudo-anomaly generator. 
\end{itemize}

\section{Related Work}
\label{sec:related}

\subsection{Overview of Anomaly Detection}
\paragraph{Reconstruction-Based Method} Reconstruction-based methods are founded on the hypothesis that models trained exclusively on normal data can successfully reconstruct normal instances while failing to do so for anomalies, thereby using reconstruction error as a measure of abnormality. Notable reconstruction-based approaches include Autoencoders (AE)\cite{2021Reconstruction, 2022Rethinking}, Generative Adversarial Networks (GANs)\cite{2014Generative, 10192551}, and Reverse Distillation (RD)\cite{Deng_2022_CVPR, Tien_2023_CVPR}. Specific methods under these categories, such as \textit{DSR}\cite{10.1007/978-3-031-19821-2_31}, \textit{RealNet}\cite{zhang2024realnetfeatureselectionnetwork}, and \textit{DRAEM}\cite{Zavrtanik_2021_ICCV}, attempt to learn the representation of normal data and generate accurate reconstructions for normal samples but tend to produce significant residuals for anomalous inputs.

\paragraph{Embedding-Based Method} Recent developments have seen the use of embedding-based methods, besides reconstruction-based methods. These models leverage pre-trained networks to extract features from input images, aiming to separate anomalies from normal samples within the feature space. Anomalies are identified based on the distance between their embeddings and those of normal samples. Techniques such as \textit{Double-MMD RGP}\cite{xiao2023restrictedgenerativeprojectiononeclass}, \textit{RealNet}\cite{zhang2024realnetfeatureselectionnetwork}, \textit{PyramidFlow}\cite{lei2023pyramidflowhighresolutiondefectcontrastive}, and \textit{CFlowAD}\cite{gudovskiy2021cflowadrealtimeunsupervisedanomaly} utilize this paradigm, some of which employ a memory bank to store features and apply distance metrics to detect anomalies. Embedding-based methods are particularly valuable in high-dimensional feature spaces, where the intricate relationships between different regions of the data can be better captured using a pre-trained model’s learned representations.

\paragraph{Synthetic Anomaly Generation} Another line of research focuses on synthetic anomaly generation, a strategy that involves creating pseudo-anomalies to augment training data and improve detection performance. These synthetic anomalies are used to provide explicit anomaly labels in the training set, converting the unsupervised task into a form of supervised learning. Notable techniques include \textit{CutPaste}~\cite{2021CutPaste}, which pastes cut-out normal patches in different positions, and methods such as \textit{GLASS}\cite{chen2024unifiedanomalysynthesisstrategy}, \textit{UniAD}\cite{you2022unifiedmodelmulticlassanomaly}, \textit{SimpleNet}\cite{Liu_2023_CVPR}, \textit{RealNet}\cite{zhang2024realnetfeatureselectionnetwork},and \textit{DDPM}~\cite{ho2020denoising}, all of which employ Gaussian noise to generate synthetic anomalies at both feature and image levels. These approaches help in training the model to differentiate between normal and abnormal samples, thus improving its ability to detect true anomalies. By contrast, methods such as \textit{DRAEM}\cite{Zavrtanik_2021_ICCV} and \textit{DMDD}~\cite{liu2024dual} employ Berlin noise or other forms of noise, which have not demonstrated the same level of effectiveness in anomaly detection tasks. In subsequent sections, we will theoretically justify the use of Gaussian noise in these pseudo-anomaly generation modules, establishing its effectiveness and reasonableness as an anomaly synthesis strategy.

\subsection{\textbf{\textit{One-class}} vs. \textbf{\textit{Multi-class}}}

Our method builds on existing frameworks in anomaly detection and requires a deep understanding of these foundational models. Without a comprehensive grasp of their structure and mechanisms, our method may be misapplied, leading to suboptimal outcomes. Hence, we will conduct demonstrative experiments from both \textit{one-class} and \textit{multi-class} perspectives.

\paragraph{\textbf{\textit{One-class}} Anomaly Detection}
\textit{One-class} anomaly detection focuses on modeling the distribution of normal data to identify deviations. Methods such as \textit{RD}~\cite{Deng_2022_CVPR} and \textit{RD++}~\cite{Tien_2023_CVPR} are foundational in this area. \textit{RD} employs a pretrain model for feature extraction, while \textit{RD++} further enhances this process for improved robustness.

\paragraph{\textbf{\textit{Multi-class}} Anomaly Detection}
\textit{Multi-class} anomaly detection handles diverse object categories and intra-class variations. Notable approaches like \textit{UniAD}~\cite{you2022unifiedmodelmulticlassanomaly} and \textit{HVQ-Trans}~\cite{lu2023hierarchical} effectively generalize across multiple classes. 

\subsection{\textbf{\textit{Non-Diffusion}} vs. \textbf{\textit{Diffusion-Based}}}

For the purpose of making our method easier for readers to apply, we have supplemented it with the diffusion-based framework.

\paragraph{Non-diffusion methods}Non-diffusion methods, such as \textit{UniAD}~\cite{you2022unifiedmodelmulticlassanomaly} and \textit{GLAD}~\cite{yao2024glad}, have been effective for anomaly detection but often face limitations in preserving the semantic consistency of anomalous regions, especially in complex settings.

\paragraph{Diffusion-based methods}Diffusion-based methods have recently gained traction due to their exceptional reconstruction capabilities, as evidenced by \textit{DiAD}~\cite{he2023diaddiffusionbasedframeworkmulticlass}, which stands out among contemporary methods. Besides, \emph{AnomalyDiffusion}~\cite{Hu2024AnomalyDiffusion} and \emph{TransFusion}~\cite{Fucka2024TransFusion}, which jointly learn to generate anomalies and discriminate them from normal data by use of Diffusion-based methods. 

\section{Method}
\subsection{Overview}

\begin{figure*}[t]
  \centering
  \resizebox{\linewidth}{!}{
  \includegraphics[width=\linewidth]{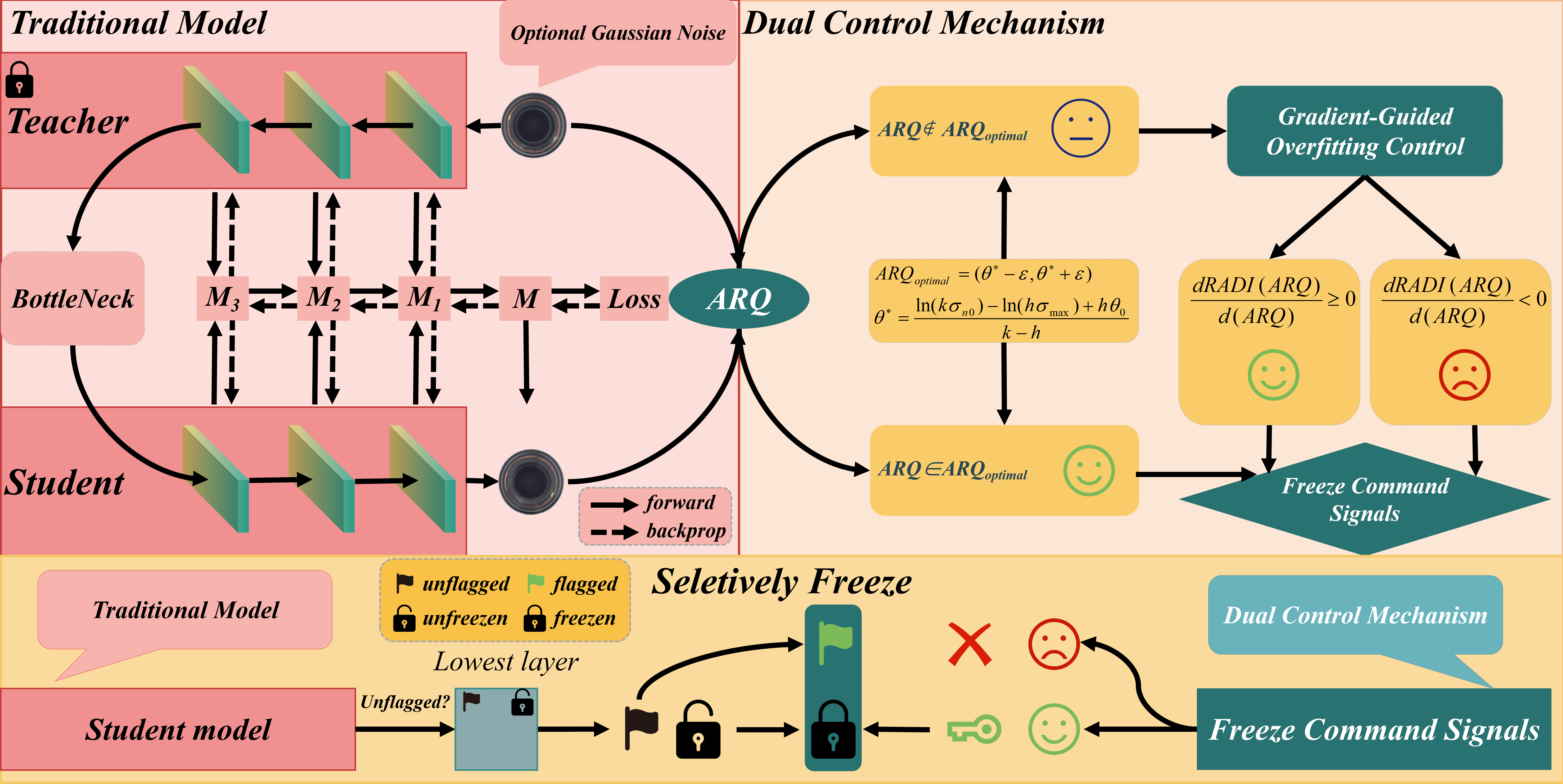}}
  % \begin{subfigure}{0.52\linewidth}  % 调整宽度比例
  %   \includegraphics[width=\linewidth]{figure/overview2_01.png}
  %   \caption{Training Stage}
  %   \label{fig:overview_train}
  % \end{subfigure}
  % \hfill
  % \begin{subfigure}{0.47\linewidth}  % 调整宽度比例
  %   \includegraphics[width=\linewidth]{figure/overview2_02.png}
  %   \caption{Inference Stage}
  %   \label{fig:overview_inference}
  % \end{subfigure}}
  \caption{Comprehensive Framework for \textit{COAD}. In the \textit{Traditional Model}, \textit{Teacher Model} remains frozen throughout. The pseudo-anomaly \textit{Gaussian Noise} is optional. The student model also remains unfrozen during normal training stage, with selective freezing employed during the controllable overfitting stages. When the \textit{Dual Control Mechanism} fails, \textit{Freeze Command Signals} are triggered, prompting the selective freezing of certain network layers and flagging the selected layer in the \textit{Traditional Model} to effectively manage overfitting. The inference pipeline follows a similar procedure to the \textit{Traditional Model}, ensuring that the enhancements provided by \textit{COAD} integrate seamlessly.}
  \label{fig:overview}
\end{figure*}

As ~\cref{fig:overview} shows, our work provides an overview of our proposed framework for enhancing anomaly detection through controllable overfitting. We introduce \textit{\textbf{C}ontrollable \textbf{O}verfitting-based \textbf{A}nomaly \textbf{D}etection}, called \textbf{\textit{COAD}}. We systematically integrate novel elements such as \textit{ARQ} for quantifying overfitting and \textit{RADI} as a complementary metric for AUROC-pixel. These components collectively facilitate the monitoring, control, and optimization of the overfitting process to maximize model sensitivity while preserving generalization capabilities. The detailed overall training and inference pipeline's pseudo code can be seen in ~\cref{alg:training_stage_detailed} and ~\cref{alg:inference_phase} in \textit{Supplementary Materials}.

Given an input image \(x \in \mathcal{X}\), our Teacher network \(\mathcal{T}\), which is trained on a large-scale dataset, produces reference features: $y \;=\; \mathcal{T}(x)$. The Student network \(\mathcal{S}\) then predicts $\hat{y} \;=\; \mathcal{S}(x)$. Finally, the discrepancy between \(y\) and \(\hat{y}\) determines the anomaly region. We get the anomaly region $  M \;=\; y \;\oplus\;\hat{y}$.
     
Besides monitoring $\text{ARQ}_{\text{optimal}} = [\,\theta - \delta,\;\theta + \delta\,]$, where \textit{ARQ} can be calculated as ~\cref{alg:arq_calculation} shown in \textit{Supplementary Materials}, we also consider a derivative-based check on $\tfrac{d\,\text{RADI}(\theta)}{d\theta} < 0$, where \textit{RADI} can be calculated as ~\cref{alg:radi_calculation} shown in \textit{Supplementary Materials}. If that condition holds, we trigger a partial freeze of the Student network by setting:$f(\mathcal{S}) \;=\; 1$. The detailed freezing process can be seen in ~\cref{alg:selective_freezing} in \textit{Supplementary Materials}.

\subsection{Quantifying Overfitting with \textbf{\textit{Aberrance Retention Quotient}}}
The exploration of overfitting has largely been confined to its avoidance, yet systematic methods to quantify, control, and beneficially leverage overfitting remain largely unexplored. To fill this gap, we introduce the \textit{\textbf{A}berrance \textbf{R}etention \textbf{Q}uotient}, called \textbf{\textit{ARQ}}, a novel metric designed to quantify the degree of overfitting by capturing the model's divergence from true data representation.

Specifically, the \textit{ARQ} is formally defined as follows:

\begin{equation}
  ARQ = \frac{\sum_{i=1}^{N} |\hat{y}_i - y_i|}{\sum_{i=1}^{N} y_i},
  \label{eq:arq-definition}
\end{equation}

\noindent where N represents the total number of instances, $\hat{y}_i$ represents the predicted output at the $i^{th}$ instance, and $y_i$ denotes the corresponding original ground truth value. The numerator captures the aggregate prediction deviation from the true labels across all data points, while the denominator represents the totality of the predicted target values, normalizing the aberrance in the context of the entire dataset.

\textit{ARQ} is a core metric in our framework that quantifies the progression of overfitting during model training. Through continuous monitoring of \textit{ARQ},  we establish an optimal interval \textit{golden overfitting interval}, where the model's sensitivity to anomalous patterns is maximized without compromising generalization. To objectively determine this interval, we employ a validation-based adaptive search strategy, optimizing \textit{ARQ} to improve anomaly detection. Specifically, we denote the interval of optimal \textit{ARQ} as:

\begin{equation}
  ARQ_{\text{optimal}} = [\theta-\delta, \theta+\delta],
  \label{eq:optimal_arq}
\end{equation}

\noindent where $\theta$ is the baseline value around which the optimal range is established, \( \delta \) is a quantity of the same order of magnitude as $\theta$, representing the permissible deviation from the baseline $\theta$ within the optimal range, which indicates the region in which the overfitting is leveraged most effectively.

As overfitting progresses, \textit{$ARQ_{optimal}$} helps us mitigate false positives while reducing false negatives, ~\cref{fig:false_positive} and ~\cref{fig:false_negative} illustrate some cases. By effectively using \textit{ARQ}, we transform overfitting into a controlled mechanism that enhances model robustness and discriminative power for anomaly detection.

\subsection{Bridging Theoretical Modeling and Practical Anomaly Detection with \textbf{\textit{Relative Anomaly Distribution Index}}}

\subsubsection{Probability Distribution Models of Normal and Anomalous Pixel Scores}
\label{sec:assumption}
To facilitate the mathematical treatment later, we make the following two assumptions. We empirically verify these assumptions and find that normal distributions provide adequate approximations for prediction score distributions, a common phenomenon justified by central limit theorem and supported by empirical observations (refer to ~\cref{sec:val_gau} and \textit{Supplementary Material}).

\paragraph{Distribution of Normal Pixel Prediction Scores}

At an \textit{Aberrance Retention Quotient} of $\textit{ARQ} = \theta$, the prediction scores of normal pixels $S_n$ follow a normal distribution dependent on $\theta$:

\begin{equation}
    S_n \sim N(\mu_n(\theta), \sigma_n(\theta)^2),
    \label{eq:dnpps}
\end{equation}

\noindent where the mean $\mu_n(\theta)$ and variance $\sigma_n(\theta)^2$ vary with \textit{ARQ}. As the \textit{ARQ} increases, the model's memory of normal samples is enhanced, potentially resulting in:

\begin{itemize}
    \item The mean $\mu_n(\theta)$ becoming closer to the average value of the training data.
    \item The variance $\sigma_n(\theta)^2$ decreasing due to the model's predictions on normal samples becoming more stable.
\end{itemize}

The variance of normal pixel prediction scores decreases exponentially with increasing \textit{ARQ}, expressed as:

\begin{equation}
    \sigma_n(\theta) = \sigma_{n0} e^{-k\theta},
    \label{eq:npps}
\end{equation}

\noindent where $\sigma_{n0}$ is the initial variance (without overfitting), and $k$ is a positive constant representing the rate at which variance decreases. This relationship shows that as the model overfits, its prediction on normal samples becomes more stable.

\paragraph{Distribution of Anomalous Pixel Prediction Scores}

The prediction scores of anomalous pixels $S_a$ follow a fixed normal distribution that does not change with \textit{ARQ}:

\begin{equation}
    S_a \sim N(\mu_a, \sigma_a^2),
    \label{eq:dapps}
\end{equation}

\noindent Since anomalous pixels represent unseen deviations that the model does not explicitly memorize during training, we assume their predicted score distribution remains relatively stable and unaffected by varying degrees of overfitting(refer to ~\cref{sec:val_gau}).

\subsubsection{Introduction of \textbf{\textit{RADI}}}

After these assumptions, we introduce the \textit{\textbf{R}elative \textbf{A}nomaly \textbf{D}istribution \textbf{I}ndex}, called \textbf{\textit{RADI}}, a novel metric designed to bridge theoretical modeling and practical anomaly detection. Unlike threshold-dependent metrics such as \textit{AUROC}, \textit{RADI} directly measures the probabilistic separation between normal and anomalous distributions, providing more explicit insights into the model's discriminative capability independent of specific thresholds. This makes it particularly valuable for theoretical analyses, while retaining practical utility in assessing model effectiveness.

The \textit{RADI} quantifies the overlap between normal and anomalous score distributions, akin to the \textit{Wilcoxon rank-sum}~\cite{wilcoxon1992individual} and \textit{Mann-Whitney U tests}~\cite{mann1947test}. \textit{RADI} is calculated using the cumulative distribution function (\textit{CDF}) of the score distributions:

\begin{equation}
    \begin{split}
        \textit{RADI}(\textit{ARQ}) &= P(S_a > S_n) \\
        &= \int_{-\infty}^{\infty} P(S_a > x) f_{S_n}(x) dx,
    \end{split}
    \label{eq:radi}
\end{equation}

\noindent where \(S_a\) and \(S_n\) represent the scores of anomalous and normal pixels, respectively. Intuitively, RADI quantifies the probability that a selected anomalous pixel has a higher anomaly score than a selected normal pixel. A higher RADI thus indicates clearer model discrimination between anomalies and normal patterns.

Although maximizing RADI might seem to encourage over-classifying anomalies, in practice, the variance of normal pixel scores remains above a non-zero lower bound due to controlled noise (~\cref{eq:noise}). This ensures normal samples retain distinct scores from anomalies, preventing the trivial solution of classifying all samples as anomalous.

\textit{RADI} provides a probabilistic assessment of whether anomalous scores exceed normal scores, offering insights into model discrimination. While similar to \textit{AUROC} in reflecting a model's ability to distinguish between normal and anomalous samples, \textit{RADI} calculates this directly through distribution comparison rather than at multiple thresholds. It is particularly useful for quick theoretical assessments, whereas \textit{AUROC} is better suited for comprehensive performance evaluations across thresholds.

Overall, \textit{RADI} complements \textit{AUROC} by providing a simplified measure of the model's discriminative power, effectively reflecting its trend without requiring threshold-based evaluations.

\subsubsection{Theoretical Derivation of Optimal Overfitting Conditions}

For our case, where both $S_a$ and $S_n$ follow normal distributions, to analytically derive the optimal level of controllable overfitting, we explicitly express RADI in closed-form under normal distribution assumptions. This facilitates understanding how statistical changes induced by overfitting influence anomaly detection performance:

\begin{equation}
    \textit{RADI}(\theta) = \Phi\left(\frac{\mu_a - \mu_n}{\sqrt{\sigma_n(\theta)^2 + \sigma_a^2}}\right),
    \label{eq:radi_normal}
\end{equation}

\noindent where $\Phi$ is the \textit{CDF} of the standard normal distribution. In this case, $\mu_a$ and $\mu_n$ are the means of the anomaly and normal pixel score distributions, respectively, while $\sigma_a$ and $\sigma_n(\theta)$ are their respective standard deviations, with $\theta$ representing the \textit{ARQ}. 

~\cref{eq:radi_normal} highlights the dependence of \textit{RADI} on the statistical properties of the score distributions. As $\theta$ increases, the standard deviation $\sigma_n(\theta)$ of the normal scores will decrease due to the model's increasing ability to memorize normal patterns, which results in a more stable distribution for normal scores. This reduction in variance directly affects the value of \textit{RADI}, increasing the probability that the anomalous scores are distinguished from normal scores, thereby enhancing the detection capability of the model.

However, when overfitting is excessive, the model may start to memorize noise, leading to instability in prediction scores, and the variance $\sigma_n(\theta)$ will no longer decrease and could even increase. To describe this phenomenon, we introduce a noise term:

\begin{equation}
\sigma_n(\theta) = \sigma_{n0} e^{-k\theta} + \sigma_{\text{noise}}(\theta),
\label{eq:noise}
\end{equation}

\noindent where $\sigma_{\text{noise}}(\theta)$ represents the noise impact induced by overfitting and can be expressed as:

\begin{equation}
\sigma_{\text{noise}}(\theta) = \sigma_{\text{max}}(1 - e^{-h(\theta - \theta_0)}), \quad \text{for} \; \theta > \theta_0,
\end{equation}

\noindent where $\sigma_{\text{max}}$ is the maximum noise variance, $h$ controls the rate of increase of noise variance, and $\theta_0$ is the threshold of \textit{ARQ} where noise starts to appear. When $\theta \leq \theta_0$, $\sigma_{\text{noise}}(\theta) = 0$.

\begin{table*}[h]
\centering
\caption{Performance comparison across different methods in \textit{one-class} tasks on \textit{MVTec AD} dataset. \textbf{Bold text} indicates the best performance among all methods, while \underline{underlined text} shows the comparison \textit{Method} \textit{vs.} \textit{Method}~(+\textit{COAD}). The values in the form of \textit{xx/xx} represent \textit{image-level AUROC / pixel-level AUROC}.}
\label{tab:one-class}
\resizebox{\textwidth}{!}{
\begin{tabular}{c|c|ccccc|cc|cc}
\toprule
 \multicolumn{2}{c|}{\textit{Category}} & \textit{DSR}~\cite{10.1007/978-3-031-19821-2_31} & \textit{PatchCore}~\cite{Roth_2022_CVPR} & \textit{BGAD}~\cite{Yao_2023_CVPR} & \textit{SimpleNet}~\cite{Liu_2023_CVPR} & \textit{GLASS}~\cite{chen2024unifiedanomalysynthesisstrategy} & \textit{RD}~\cite{Deng_2022_CVPR} & \textit{RD} (+\textit{COAD}) & \textit{RD++}~\cite{Tien_2023_CVPR} & \textit{RD++} (+\textit{COAD}) \\
\midrule
\multirow{10}{*}{\rotatebox{90}{\textit{Object}}} 
 &\textit{Bottle}     & 99.6/98.8 & \textbf{100}/98.5 & \textbf{100}/98.9 & \textbf{100}/98.0 & \textbf{100}/99.3 & \textbf{100}/98.7 & \underline{\textbf{100}/98.8} & \textbf{100}/98.8 & \underline{\textbf{100/99.6}} \\
 &\textit{Cable}      & 95.3/97.7 & 99.8/98.4 & 97.9/98.0 & \textbf{100}/97.5 & 99.8/98.7 & 95.0/97.3 & \underline{97.9/97.7} & 99.3/98.4 & \underline{\textbf{99.4}/\textbf{99.2}} \\
 &\textit{Capsule}    & 98.3/91.0 & 98.1/99.0 & 97.3/98.0 & 97.8/98.9 & \textbf{99.9}/99.4 & 96.3/98.3 & \underline{98.8/98.4} & 99.0/98.8 & \underline{99.7/\textbf{99.5}} \\
 &\textit{Hazelnut}   & 97.7/99.1 & \textbf{100}/98.7 & 99.3/98.5 & 99.8/98.1 & \textbf{100}/\textbf{99.4} & 99.9/98.9 & \underline{\textbf{100}/99.1} & \textbf{100}/99.2 & \underline{\textbf{100/99.4}} \\
 &\textit{Metal Nut}   & 99.1/94.1 & \textbf{100}/98.3 & 99.3/97.7 & \textbf{100}/98.8 & \textbf{100}/99.4 & \textbf{100}/97.3 & \underline{\textbf{100}/97.7} & \textbf{100}/98.1 & \underline{\textbf{100/99.6}} \\
 &\textit{Pill}       & 98.9/94.2 & 96.4/97.8 & 98.8/98.0 & 98.6/98.6 & 99.3/\textbf{99.4} & 96.6/98.2 & \underline{97.9/98.4} & 98.4/98.3 & \underline{\textbf{99.5}/99.0} \\
 &\textit{Screw}      & 95.9/98.1 & 98.4/99.5 & 92.3/99.2 & 98.7/99.2 & \textbf{100}/99.5 & 97.0/99.5 & \underline{99.5/99.6} & 98.9/\textbf{\underline{99.7}} & \underline{\textbf{99.9}}/99.6 \\
 &\textit{Toothbrush} & \textbf{100}/\textbf{99.5} & \textbf{100}/98.6 & 86.9/98.7 & \textbf{100}/98.5 & \textbf{100}/99.3 & 99.5/98.9 & \underline{\textbf{100}/99.1} & \textbf{100}/99.1 & \underline{\textbf{100}/99.4} \\
 &\textit{Transistor} & 96.3/80.3 & 99.9/96.1 & 99.7/93.9 & \textbf{100}/97.0 & 99.9/97.6 & 96.7/92.4 & \underline{98.9/93.4} & 98.5/94.3 & \underline{99.7/\textbf{98.9}} \\
 &\textit{Zipper}     & 98.5/98.4 & 99.4/98.9 & 97.8/98.7 & 99.9/98.9 & \textbf{100}/\textbf{99.6} & 98.5/98.2 & \underline{99.1/98.7} & 98.6/98.8 & \underline{\textbf{99.8}/99.5} \\
\midrule
\multirow{5}{*}{\rotatebox{90}{\textit{Texture}}} 
 &\textit{Carpet}     & 99.6/96.0 & 98.6/99.1 & 99.8/99.4 & 99.7/98.4 & 99.8/\textbf{99.6} & 98.9/98.9 & 99.8/99.2 & \underline{\textbf{100}}/99.2 & 99.8/\underline{99.5} \\
 &\textit{Grid}       & \textbf{100}/\textbf{99.6} & 97.7/98.8 & 99.1/99.4 & 99.9/98.5 & \textbf{100}/99.4 & \textbf{100}/99.1 & \underline{\textbf{100}/99.3} & \textbf{100}/99.3 & \underline{\textbf{100}/99.5} \\
 &\textit{Leather}    & 99.3/99.5 & \textbf{100}/99.3 & \textbf{100}/99.7 & \textbf{100}/99.2 & \textbf{100}/99.8 & \textbf{100}/99.4 & \underline{\textbf{100}/99.5} & \textbf{100}/99.5 & \underline{\textbf{100}/\textbf{99.9}} \\
 &\textit{Tile}       & \textbf{100}/98.6 & 98.8/95.7 & \textbf{100}/96.7 & 98.7/97.7 & \textbf{100}/\textbf{99.7} & 99.3/95.6 & \underline{99.9/95.7} & 99.7/96.6 & \underline{\textbf{100}/98.8} \\
 &\textit{Wood}       & 94.7/91.5 & 99.1/95.0 & 99.5/97.0 & 99.5/94.4 & \textbf{99.9}/\textbf{98.8} & 99.2/95.3 & \underline{99.6/95.5} & 99.3/95.8 & \underline{\textbf{99.9/98.8}} \\
\midrule
\rowcolor[gray]{0.9}
 \multicolumn{2}{c|}{\textit{\textbf{Avg.}}}   & 98.2/95.8 & 99.1/98.1 & 97.9/98.2 & 99.5/98.1 & \textbf{99.9}/99.3 & 98.5/97.7 & \underline{99.4/98.0} & 99.4/98.3 & \underline{\textbf{99.9/99.4}} \\
\bottomrule
\end{tabular}
}
\end{table*}

\begin{table*}[htbp]
\centering
\caption{Performance comparison across different methods in \textit{one-class} tasks on \textit{VisA} dataset.}
\label{tab:one-class-visa}
\resizebox{\textwidth}{!}{
\begin{tabular}{c|c|ccc|cc|cc}
\toprule
 \multicolumn{2}{c|}{\textit{Category}} & \textit{PatchCore}~\cite{Roth_2022_CVPR} & \textit{SimpleNet}~\cite{Liu_2023_CVPR} & \textit{DRAEM}~\cite{Zavrtanik_2021_ICCV}  & \textit{RD}~\cite{Deng_2022_CVPR} 
&  \textit{RD}(+\textit{COAD})&\textit{RD++}~\cite{Tien_2023_CVPR}&\textit{RD++}(+\textit{COAD})\\ \midrule
\multirow{4}{*}{\textit{Single Instance}}
 &\textit{Cashew}            & 97.7/\textbf{99.2}                   & 94.8/99.0          & 88.3/85.0                 & \underline{\textbf{99.6}}/95.8 &  97.8/\underline{97.6}         
&97.8/95.8           &\underline{98.1}/\underline{98.3}         
\\
 &\textit{Chewinggum}        & 99.1/98.9                   & \textbf{100}/98.5           & 96.4/97.7                 & \underline{99.7}/99.0 &  97.8/\underline{99.4}         
&96.4/99.4        &\underline{97.4}/\underline{\textbf{99.5}}
\\
 &\textit{Fryum}             & 91.6/95.9                   & 96.6/94.5                   & 94.7/82.5                 & \underline{96.6}/94.3          &  95.5/\underline{96.3}         
&95.8/96.5           &\underline{\textbf{96.8}}/\underline{\textbf{98.1}} 
\\
 &\textit{Pipe fryum}        & 99.0/99.3                   & 99.2/99.3                   & 94.7/65.8                 & 97.0/92.0          &  \underline{99.4}/\underline{99.1}         
&\underline{\textbf{99.6}}/99.1 &99.3/\underline{\textbf{99.4}}
\\ 
\midrule
\multirow{4}{*}{\textit{Multiple Instances}}
 &\textit{Candle}            & \textbf{98.7}/\textbf{99.2}          & 96.9/98.6                   & 89.6/91.0                 & 92.2/97.9           &  \underline{95.1}/\underline{99.0}          
&\underline{96.4}/98.6  &92.8/\underline{98.7}         
\\
 &\textit{Capsules}          & 68.8/96.5                   & 89.5/99.2                   & 89.2/99.0                 & \underline{90.1}/89.5          &  87.5/\underline{99.3}         
&92.1/99.4 &\underline{\textbf{92.7}}/\underline{\textbf{99.7}} 
\\
 &\textit{Macaroni1}         & 90.1/98.5                   & 97.6/99.6                   & 93.9/99.4                 & \underline{\textbf{98.4}}/97.7 &  95.1/\underline{99.3}         
&94.0/99.7 &\underline{95.7}/\underline{\textbf{99.9}}
\\
 &\textit{Macaroni2}         & 63.4/93.5                   & 83.4/96.4                   & 88.3/\textbf{99.7}                 & \underline{\textbf{97.6}}/87.7 &  96.6/\underline{99.3}         
&88.0/\underline{\textbf{99.7}} &\underline{92.5}/99.6         
\\
\midrule
\multirow{4}{*}{\textit{Complex Structure}}
 &\textit{PCB1}              & 96.0/\textbf{99.8}          & \textbf{99.2}/\textbf{99.8}          & 84.7/98.4                 & \underline{97.6}/75.0          &  96.5/\underline{99.7}
&97.0/\underline{99.7} &\underline{98.8}/99.5 
\\
 &\textit{PCB2}              & 95.1/98.4                   & \textbf{99.2}/98.8          & 96.2/94.0                 & 91.1/64.8          &  \underline{94.0}/\underline{98.8}         
&97.2/\underline{\textbf{99.0}} &\underline{97.3}/98.9 
\\
 &\textit{PCB3}              & 93.0/98.9                   & \textbf{98.6}/99.2                   & 97.4/94.3                 & 95.5/95.5          &  \underline{95.9}/\underline{\textbf{99.3}}         
&96.8/99.2          &\underline{97.4}/\underline{\textbf{99.3}}
\\
 &\textit{PCB4}              & 99.5/98.3                   & 98.9/98.6                   & 98.8/97.6                 & 96.5/92.8          &  \underline{\textbf{99.9}}/\underline{98.7}
&\underline{99.8}/98.6 &98.5/\underline{\textbf{99.6}}
\\
\midrule
\rowcolor[gray]{0.9}
 \multicolumn{2}{c|}{\textit{\textbf{Avg.}}}           & 91.0/98.1                   & 96.2/98.5                   & 92.4/92.0                 & \underline{96.0}/90.1          &  95.9/\underline{98.8}         &95.9/98.7          &\underline{\textbf{96.4}}/\underline{\textbf{99.2}}\\ \bottomrule
\end{tabular}}
\end{table*}

To find the optimal \textit{Aberrance Retention Quotient} of $\textit{ARQ} = \theta^*$ that maximizes \textit{RADI}, we calculate the derivative of $\text{AUROC}_{\text{pixel}}(\theta)$ and set it to zero.

The derivative is as below and we set the derivative to zero:

\begin{equation}
    \frac{d\textit{RADI}(\theta)}{d\theta} = \phi(z(\theta)) \cdot \frac{d z(\theta)}{d \theta} = 0,
    \label{eq:derivative}
\end{equation}

\noindent where $\phi(z)$ is the probability density function of the standard normal distribution.

\begin{equation}
    \frac{d z(\theta)}{d \theta} = - \frac{(\mu_a - \mu_n) \cdot \sigma_n(\theta) \cdot \frac{d \sigma_n(\theta)}{d \theta}}{(\sigma_n(\theta)^2 + \sigma_a^2)^{3/2}} = 0,
    \label{eq:derivative_process}
\end{equation}

\noindent This implies either $\sigma_n(\theta) = 0$ (unrealistic as variance cannot be zero) or $\frac{d \sigma_n(\theta)}{d \theta} = 0$. Thus, solving yields:

\begin{equation}
    \theta^* = \frac{\ln(k\sigma_{n0}) - \ln(h\sigma_{\text{max}}) + h\theta_0}{k - h}.
    \label{eq:derivative_result}
\end{equation}

\subsubsection{\textbf{\textit{Dual Control Mechanism}} and \textbf{\textit{Freeze Command Signals}}}

We define \textit{Aberrance Retention Quotient} of $\textit{ARQ} = \theta$, adjusted to optimize the \textit{pixel-level AUROC}. The mathematical relationship between \textit{ARQ} and \textit{AUROC} is established by monitoring the gradient, we call it \textit{Gradient-Guided Overfitting Control}. The \textit{Dual Control Mechanism} leverages both \textit{ARQ} and \textit{RADI} simultaneously: \textit{ARQ} ensures overfitting is maintained within beneficial bounds, while \textit{RADI}’s gradient condition ensures the continual improvement of anomaly detection performance.

\begin{equation}
    \frac{d\textit{RADI}(\theta)}{d\theta} \geq 0.
    \label{eq:control_overfit}
\end{equation}

\noindent We integrate \textit{$ARQ_{optimal}$} in ~\cref{eq:optimal_arq} and \textit{Gradient-Guided Overfitting Control} in ~\cref{eq:control_overfit} and call this combined control mechanism the \textit{Dual Control Mechanism}:

\begin{equation}
    \frac{d\textit{RADI}(\theta)}{d\theta} \geq 0, \quad \text{for} \; \theta \in \textit{$ARQ_{optimal}$}.
    \label{eq:dual_control_overfit}
\end{equation}

\noindent When \textit{ARQ} exceeds its optimal bounds and \textit{RADI} gradient condition fails, \textit{Freeze Command Signals} are activated, progressively freezing layers starting from the earliest layers upwards.

The rationale for progressively freezing certain layers while overfitting the other parts lies in the architecture's functional partitioning. Lower layers are responsible for extracting fundamental features, which are common across both normal and anomalous data. Freezing these layers early preserves their ability to extract generalizable features, preventing degradation due to overfitting. Conversely, keeping deeper layers unfrozen during the overfitting stage enables the model to specialize in identifying unique, high-level features that are more discriminative between normal and anomalous samples. This approach ultimately boosts performance without succumbing to the pitfalls of overfitting the entire model.

\section{Experiments}
\label{sec:exp}

\begin{table*}[h]
\centering
\caption{Performance comparison across different methods in \textit{multi-class} tasks on \textit{MVTec AD} dataset.}
\label{tab:multi-class}
\resizebox{\textwidth}{!}{
\begin{tabular}{c|c|ccccc|cc|cc}
\toprule
 \multicolumn{2}{c|}{\textit{Category}}& \textit{US}~\cite{Bergmann_2020_CVPR}    & \textit{PSVDD}~\cite{Yi_2020_ACCV} & \textit{PaDiM}~\cite{10.1007/978-3-030-68799-1_35}  & \textit{MKD}~\cite{Salehi_2021_CVPR}   & \textit{DRAEM}~\cite{Zavrtanik_2021_ICCV} & \textit{UniAD}~\cite{you2022unifiedmodelmulticlassanomaly} & \textit{UniAD} (+\textit{COAD}) & \textit{DiAD}~\cite{he2023diaddiffusionbasedframeworkmulticlass} & \textit{DiAD} (+\textit{COAD})\\
\midrule
\multirow{10}{*}{\rotatebox{90}{\textit{Object}}} 
 &\textit{Bottle}      & 84.0/67.9  & 85.5/86.7  & 97.9/96.1   & 98.7/91.8  & 97.5/87.6  & 99.7/\underline{98.1}  & \underline{\textbf{100}}/98.0 & 99.7/98.4 & \underline{\textbf{100}/\textbf{98.8}} \\
 &\textit{Cable}       & 60.0/78.3  & 64.4/62.2  & 70.9/81.0   & 78.2/89.3  & 57.8/71.3  & 95.2/97.3  & \underline{\textbf{97.9}/\textbf{98.8}} & 94.8/96.8 & \underline{97.5/97.7}\\
 &\textit{Capsule}     & 57.6/85.5  & 61.3/83.1  & 73.4/96.9   & 68.3/88.3  & 65.3/50.5  & 86.9/98.5  & \underline{92.3/98.6} & 89.0/97.1 & \underline{\textbf{98.0}/\textbf{98.8}}\\
 &\textit{Hazelnut}    & 95.8/93.7  & 83.9/97.4  & 85.5/96.3   & 97.1/91.2  & 93.7/96.9  & 99.8/\underline{98.1}  & \underline{99.9/98.1} & 99.5/98.3 & \underline{\textbf{100}/\textbf{99.1}} \\
 &\textit{Metal Nut}   & 62.7/76.6  & 80.9/96.0  & 88.0/84.8   & 64.9/64.2  & 72.8/62.2  & 99.2/\underline{94.8}  & \underline{99.3}/93.2 & 99.1/97.3 & \underline{\textbf{100/97.7}} \\
 &\textit{Pill}        & 56.1/80.3  & 89.4/96.5  & 68.8/87.7    & 79.7/69.7  & 82.2/94.4  & 93.7/95.0  & \underline{94.4/\textbf{98.4}} & 95.7/95.7 & \underline{\textbf{97.4}/\textbf{98.4}}\\
 &\textit{Screw}       & 66.9/90.8  & 80.9/74.3  & 56.9/94.1   & 75.6/92.1  & 92.0/95.5  & 87.5/98.3  & \underline{90.6/98.6} & 90.7/97.9 & \underline{\textbf{99.5}/\textbf{99.6}}\\
 &\textit{Toothbrush}  & 57.8/86.9  & 99.4/98.0  & 95.3/95.6   & 75.3/88.9  & 90.6/97.7  & 94.2/98.4  & \underline{94.7/98.6} & 99.7/99.0 & \underline{\textbf{100}/\textbf{99.1}}\\
 &\textit{Transistor}  & 61.0/68.3  & 77.5/78.5  & 86.6/92.3    & 73.4/71.7  & 74.8/64.5  & \underline{\textbf{99.8}}/97.9  & \underline{\textbf{99.8}/98.4} & \underline{\textbf{99.8}/95.1} & \underline{\textbf{99.8}}/93.4\\
 &\textit{Zipper}      & 78.6/84.2  & 77.8/95.1  & 79.7/94.8    & 87.4/86.1 & \textbf{98.8}/98.3  & 95.8/96.8  & \underline{97.6/98.1} & 95.1/96.2 & \underline{98.3/\textbf{98.7}}\\
\midrule
\multirow{5}{*}{\rotatebox{90}{\textit{Texture}}} 
 &\textit{Carpet}      & 86.6/88.7  & 63.3/78.6  & 93.8/97.6    & 69.8/95.5  & 98.0/98.6  & 99.8/98.5  & \underline{\textbf{100}/98.8} & 99.4/98.6 & \underline{99.6/\textbf{99.2}}\\
 &\textit{Grid}        & 69.2/64.5  & 66.0/70.8  & 73.9/71.0    & 83.8/82.3  & 99.3/98.7  & 98.2/96.5  & \underline{99.1/98.4} & 98.5/96.6 & \underline{\textbf{100}/\textbf{99.3}}\\
 &\textit{Leather}     & 97.2/95.4  & 60.8/93.5  & 99.9/84.8    & 93.6/96.7  & 98.7/97.3  & \underline{\textbf{100}/98.8}  & \underline{\textbf{100}/98.8} & 99.8/98.8 & \underline{\textbf{100}/\textbf{99.5}}\\
 &\textit{Tile}        & 93.7/82.7  & 88.3/92.1  & 93.3/80.5   & 89.5/85.3  & \textbf{99.8}/98.0  & \underline{99.3}/91.8  & \underline{99.3/\textbf{98.6}} & 96.8/92.4 & \underline{99.3/95.7}\\
 &\textit{Wood}        & 90.6/83.3  & 72.1/80.7  & 98.4/89.1    & 93.4/80.5  & \textbf{99.8}/96.0  & 98.6/93.2  & \underline{98.5/\textbf{96.1}} & 99.7/93.3 & \underline{98.7/95.5} \\
\midrule
\rowcolor[gray]{0.9}
\multicolumn{2}{c|}{\textbf{\textit{Avg.}}}   & 74.5/81.8  & 76.8/85.6  & 84.2/89.5    & 81.9/84.9  & 88.1/87.2  & 96.5/96.8  & \underline{97.6/\textbf{98.0}} & 97.2/96.8 & \underline{\textbf{99.1}/\textbf{98.0}}\\
\bottomrule
\end{tabular}
}
\end{table*}

% \subsection{Dataset}

We evaluate our method using the \textit{MVTec AD} dataset~\cite{bergmann2019mvtec} and \textit{VisA} dataset~\cite{zou2022spot}, which are comprehensive, real-world datasets for unsupervised anomaly detection tasks. The \textit{MVTec AD} dataset includes 15 categories, comprising 3629 images for training and 1725 images for testing. The \textit{VisA} dataset includes 12 categories, including 9,621 normal and 1,200 anomalous samples.

\subsection{Comparison with SOTAs}

In this section, we compare the performance of our proposed method with several SOTA anomaly detection methods in both \textit{one-class} and \textit{multi-class} tasks on \textit{MVTec~AD} and \textit{VisA} datasets. 

Using the optimal \textit{ARQ} value $\theta$ derived from the \textit{$ARQ_{optimal}$} as indicated by ~\cref{eq:derivative_result}, we set \textit{ARQ} to 0.006 for \textit{one-class} tasks and 0.06 for \textit{multi-class} tasks as shown in ~\cref{sec:arq_ablation} which is chosen empirically. \textit{$ARQ_{optimal}$} is defined as \(\theta \in [0.001, 0.011]\) for \textit{one-class} tasks and \(\theta \in [0.01, 0.11]\) for \textit{multi-class} tasks, where the balance between overfitting and generalization is optimal for anomaly detection.

Furthermore, during model training, we adopt a \textit{Dual Control Mechanism} for Optimal Performance method in our framework \textit{COAD}, as detailed in ~\cref{eq:dual_control_overfit}, and when \textit{Dual Control Mechanism} fails, it triggers \textit{Freeze Command Signals}, prompting selective freezing of certain network layers in the traditional
framework.

Our method builds upon existing anomaly detection frameworks. Initially, we perform standard training on the original model. Following this, we transition into the controllable overfitting stage, during which the model is further trained using our \textit{COAD} framework. This stage utilizes \textit{Dual Control Mechanism} in ~\cref{eq:dual_control_overfit}. All pseudo-anomaly generation in this stage is performed using \textit{Gaussian noise} to introduce controlled variability, which aids in refining the model's sensitivity to anomalies. Additionally, the learning rate is reduced to $\frac{1}{10}$ of that used in the normal training stage, allowing the model to make finer adjustments.

After training, the inference of the model proceeds in the same manner as the baseline anomaly detection frameworks, ensuring that the enhancements provided by \textit{COAD} integrate seamlessly.

\begin{figure}[h] 
\centering 
\includegraphics[width=\linewidth]{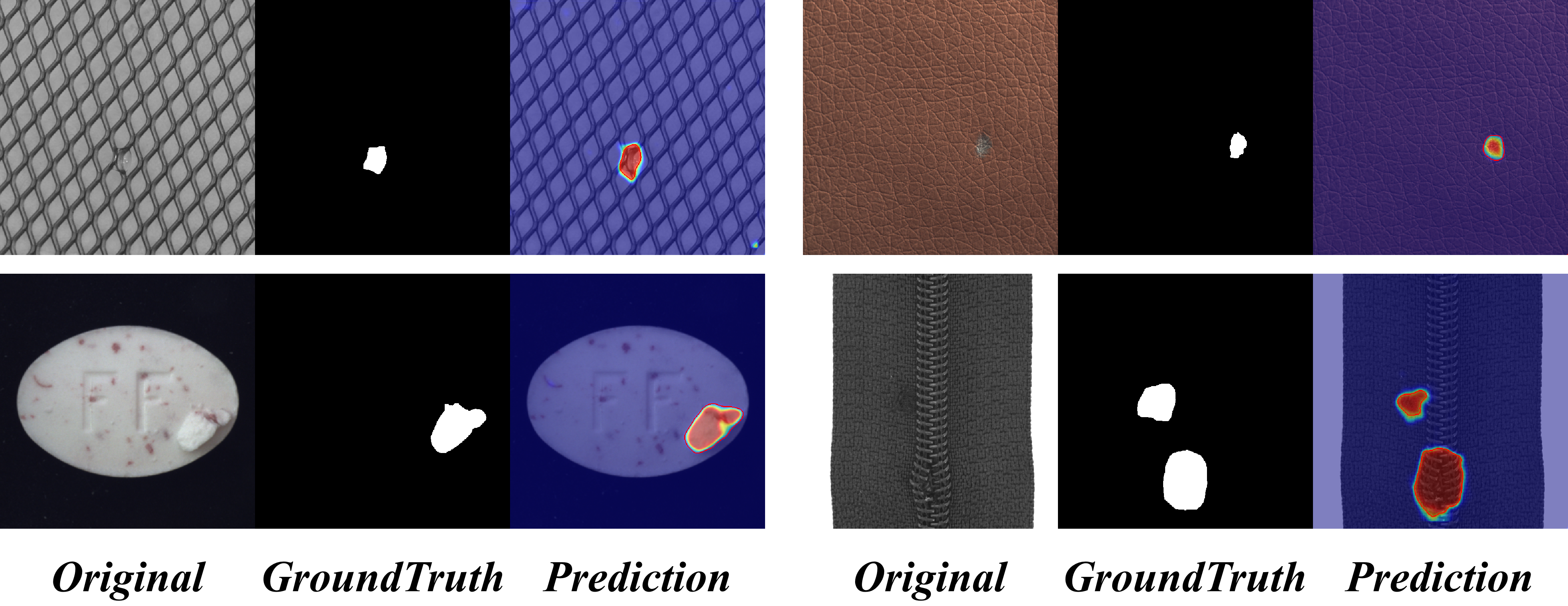} 
\caption{Qualitative visualization results on partial categories from the \textit{MVTec~AD} for the \textit{one-class} task. Columns show \textit{Original} images, \textit{GroundTruth}, and predictions generated by our \textit{COAD} framework.} 
\label{fig:partial_oneclass} 
\end{figure}

\begin{figure}[h] 
\centering 
\includegraphics[width=\linewidth]{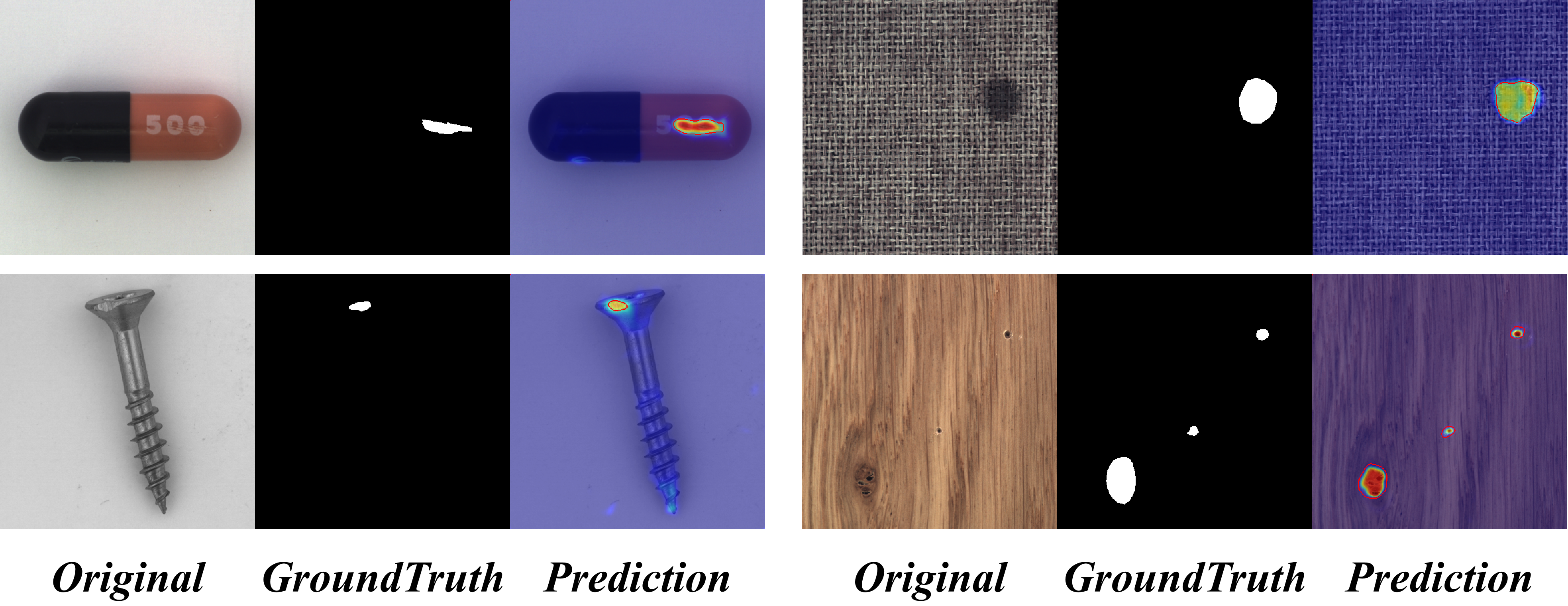} 
\caption{Qualitative visualization results on partial categories from the \textit{MVTec~AD} for the \textit{multi-class} task.} 
\label{fig:partial_multiclass} 
\end{figure}

The quantitative results of our experiments across both \textit{one-class} and \textit{multi-class} anomaly detection tasks are comprehensively summarized in ~\cref{tab:one-class,tab:one-class-visa,tab:multi-class}, with additional detailed results for the \textit{VisA} \textit{multi-class} task provided in ~\cref{tab:multi-class-visa} of the \textit{Supplementary Materials} ~\cref{sec:additional_visa}. Clearly, our proposed \textit{COAD} framework consistently surpasses previous \textit{SOTA} approaches, delivering substantial improvements in both \textit{image-level} and \textit{pixel-level AUROC} metrics across all evaluated datasets and tasks. Additionally, partial qualitative visualization results demonstrating the effectiveness of \textit{COAD} are presented in ~\cref{fig:partial_oneclass,fig:partial_multiclass}, showing high accuracy. Complete qualitative comparisons are provided in ~\cref{sec:qualitativecomparison} of the \textit{Supplementary Materials}.

In conclusion, these comprehensive experimental validations solidify the practical advantages of our \textit{COAD} framework, highlighting its ability to reliably enhance existing anomaly detection models through controlled and strategic utilization of overfitting.

\subsection{Ablation Study}

\subsubsection{Comparisons with Our Basic Frameworks}

\begin{table}[h]
        \caption{Image-Level AUROC comparison between the basic and enhanced versions of different frameworks.}
    \label{tab:auroc_image}
    \centering
    \resizebox{\linewidth}{!}{
    \begin{tabular}{clcccc}
        \toprule
         \multicolumn{2}{c}{}& \textit{RD}       & \textit{RD++}      & \textit{UniAD}      & \textit{DiAD}       \\
        \midrule
         \multirow{2}{*}{\textit{MVTec~AD}}&\textit{Baseline}       & 98.5     & 99.4      & 96.5       & 97.2       \\
         &+\textit{COAD}       & 99.4(\textcolor{red}{+\textit{0.9}})  & 99.9(\textcolor{red}{+\textit{0.5}}) & 97.6(\textcolor{red}{+\textit{1.1}}) & 99.1(\textcolor{red}{+\textit{1.9}}) \\
        \midrule
         \multirow{2}{*}{\textit{VisA}}& \textit{Baseline}       
         & 96.0& 95.9& 85.5&-\\
         & +\textit{COAD}       & 95.9(\textcolor{blue}{-\textit{0.1}})& 96.4(\textcolor{red}{+\textit{0.5}})& 91.0(\textcolor{red}{+\textit{5.5}})  &-\\
         \bottomrule
    \end{tabular}}

\end{table}

\begin{table}[h]
    \centering
    \caption{Pixel-Level AUROC comparison between the basic and enhanced versions of different frameworks.}
    \label{tab:auroc_pixel}
    \resizebox{\linewidth}{!}{
    \begin{tabular}{clcccc}
        \toprule
         \multicolumn{2}{c}{}& \textit{RD}       & \textit{RD++}      & \textit{UniAD}      & \textit{DiAD}       \\
        \midrule
         \multirow{2}{*}{\textit{MVTec~AD}}&\textit{Baseline}       & 97.7     & 98.3      & 96.8       & 96.8       \\
         &+\textit{COAD}       & 98.0(\textcolor{red}{+\textit{0.3}})  & 99.4(\textcolor{red}{+\textit{1.1}}) & 98.0(\textcolor{red}{+\textit{1.2}}) & 98.0(\textcolor{red}{+\textit{1.2}}) \\
        \midrule
         \multirow{2}{*}{\textit{VisA}}& \textit{Baseline}       
&90.1 &98.7 &95.9 &-\\
         & +\textit{COAD}       &98.8(\textcolor{red}{+\textit{8.7}}) &99.2(\textcolor{red}{+\textit{0.5}}) &97.5(\textcolor{red}{+\textit{1.6}}) &-\\
        \bottomrule
    \end{tabular}}
\end{table}

We conduct an ablation study comparing our enhanced framework with its corresponding basic versions in ~\cref{tab:auroc_image,tab:auroc_pixel}. The results demonstrate consistent and notable improvements in both \textit{image-level} and \textit{pixel-level AUROC} across all evaluated frameworks. 

In particular, the performance gains observed in \textit{image-level} anomaly detection emphasize the strength of our \textit{COAD}, which effectively captures subtle differences between normal and anomalous regions. Specifically, our enhancements yield an increase of up to \textit{1.9} in the \textit{DiAD} framework, as seen in \textit{image-level AUROC}, and similar significant improvements in other frameworks.

For \textit{pixel-level AUROC}, a similar pattern of improvement is observed, with gains of up to \textit{1.2} seen in multiple frameworks, as shown in \cref{tab:auroc_pixel}. This underscores the strength of our \textit{COAD}, which not only optimizes image-level detection but also provides enhanced granularity in distinguishing pixel-level abnormalities.

Furthermore, the versatility of our method is evident in its applicability to different anomaly detection paradigms. Our method has consistently demonstrated superior performance, showcasing our method's robustness and generalizability.

\subsubsection{Comparisons with Distinct \textit{ARQ}}
\label{sec:arq_ablation}

\begin{table}[h]
\centering
\caption{Comparison of performance across distinct \textit{ARQ} ranges for different models in \textit{one-class} tasks on \textit{MVTec~AD}.}
\label{tab:arq_one_comparison}
\resizebox{\linewidth}{!}{
\begin{tabular}{lccc}
\toprule
\textit{Model} & 0--0.001 & 0.001--0.011 & 0.011--0.1 \\
\midrule
\textit{RD} & 98.8 / 97.8 & 99.4 / 98.0 & 99.6 / 97.6 \\
\textit{RD++} & 99.7 / 98.8 & 99.9 / 99.4 & - \\
\bottomrule
\end{tabular}}
\end{table}

\begin{table}[h]
\centering
\caption{Comparison of performance across distinct \textit{ARQ} ranges for different models in \textit{multi-class} tasks on \textit{MVTec~AD}.}
\label{tab:arq_mul_comparison}
\resizebox{\linewidth}{!}{
\begin{tabular}{lccc}
\toprule
\textit{Model} & 0--0.01 & 0.01--0.11 & 0.11--1 \\
\midrule
\textit{UniAD} & 97.0 / 96.8 & 97.6 / 98.0 & 98.1 / 96.9 \\
\textit{DiAD} & 97.3 / 96.6 & 99.1 / 98.0 & 99.2 / 96.5 \\
\bottomrule
\end{tabular}}
\end{table}

In this section, we compare the performance of different models under distinct \textit{ARQ} ranges to validate the impact of controllable overfitting on anomaly detection. 

We evaluate methods such as \textit{RD}, \textit{RD++}, \textit{UniAD}, and \textit{DiAD} across various \textit{ARQ} levels. As ~\cref{tab:arq_one_comparison,tab:arq_mul_comparison} show, the results demonstrate that, by appropriately regulating the \textit{ARQ}, which is in \textit{$ARQ_{optimal}$}, model detection can be significantly enhanced. It is evident that our controllable overfitting strategy not only yields consistent improvements across different models but also helps in striking the balance between overfitting and generalization.

Overall, these results demonstrate that our controllable overfitting strategy consistently yields significant performance gains across various models and tasks. By leveraging the \textit{ARQ} within the identified optimal range, we successfully enhance the models' discriminative abilities while maintaining a balanced relationship between overfitting and generalization.

\subsection{Validation of Distribution Assumptions and Theoretical Validation of \textbf{\textit{Gaussian Noise}} Usage}

\begin{figure}[h]
    \centering
    % Normal 子图部分
    \resizebox{\linewidth}{!}{
    \begin{tabular}{cc}
        \subcaptionbox{Carpet (Normal)}{
            \includegraphics[width=0.5\textwidth]{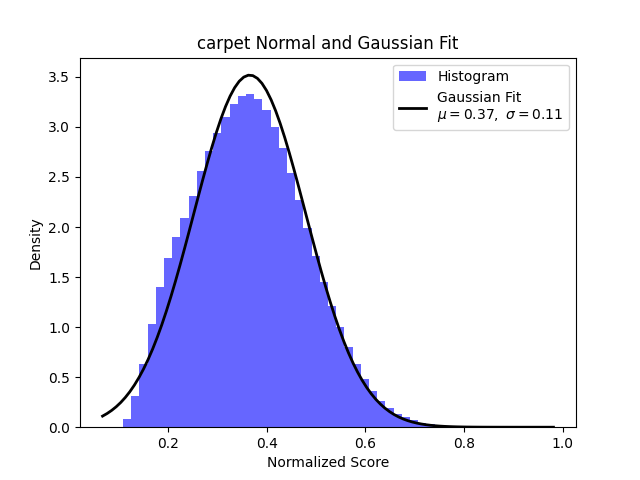}
        } &
        \subcaptionbox{Wood (Normal)}{
            \includegraphics[width=0.5\textwidth]{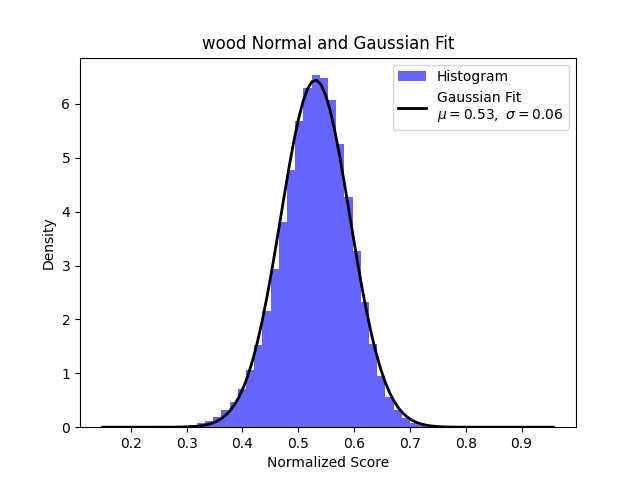}
        } \\
                \subcaptionbox{Carpet (Anomaly)}{
            \includegraphics[width=0.5\textwidth]{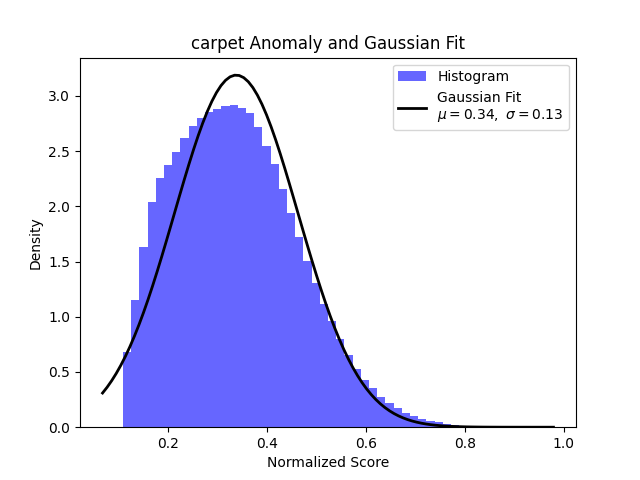}
        } &
        \subcaptionbox{Wood (Anomaly)}{
            \includegraphics[width=0.5\textwidth]{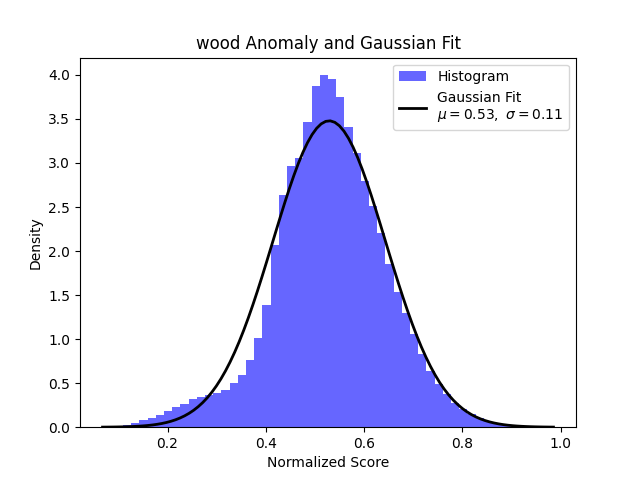}
        } \\
    \end{tabular}}
    \caption{Histograms for normal and anomaly pixel scores of partial categories of \textit{MVTec~AD} dataset. Each histogram includes Gaussian fitting curves for better visualization.}
    \label{fig:partial_scores}
\end{figure}

\label{sec:val_gau}
\begin{table}[h]
  \centering
  \caption{Partial Results of Validation Metrics for Assumptions on Normal and Anomalous Pixel Score Distributions.}
  \label{tab:tvd_results}
  \resizebox{\linewidth}{!}{
  \begin{tabular}{@{}lcccccc@{}}
    \toprule
    \textit{ARQ} & $\mu_{n}$ & $\theta_{n}$ & $\mu_{a}$ & $\theta_{a}$ & $TVD_{n}$ & $TVD_{a}
$\\
    \midrule
    0 & 105.6 & 65.5 & 99.3 & 45.0 & 0.20 & 0.08 \\
    0.006 & 105.3 & 65.5 & 98.9 & 44.9 & 0.20 & 0.08 \\
    0.015 & 105.6 & 65.5 & 99.0 & 45.1 & 0.20 & 0.08\\
    0.045 & 105.8 & 65.4 & 99.3 & 45.1 & 0.20 & 0.08 \\
    0.050 & 105.6 & 65.3 & 99.1 & 45.0 & 0.21 & 0.09 \\
    0.060 & 105.6 & 65.2 & 99.0 & 45.0 & 0.20 & 0.09 \\
    0.082 & 105.5 & 65.0 & 99.0 & 45.0 & 0.19 & 0.08 \\
    \bottomrule
  \end{tabular}}
\end{table}

In this section, we firstly validate the assumptions introduced in~\cref{sec:assumption} by employing the Total Variation Distance (TVD) metric. We leverage multiple models at different stages of overfitting, characterized by distinct \textit{ARQ} values. For each model state, corresponding to different \textit{ARQ} = \(\theta\), we calculate the prediction scores for all pixels across the dataset. Subsequently, we evaluate their proximity to a \textit{Gaussian} distribution using \textit{TVD}. The partial results of this validation are summarized in~\cref{tab:tvd_results}. Moreover, we model the variance of the normal pixel prediction scores \((\sigma_n(\theta))\) as an exponential function of the \textit{ARQ}, given by~\cref{eq:npps}. Partial qualitative result can be seen in ~\cref{fig:partial_scores} and complete qualitative study can be seen in ~\cref{fig:normal_histograms,fig:anomaly_histograms} in \textit{Supplementary Materials}.

To validate the suitability of Gaussian noise as a pseudo-anomaly generator, we employ the TVD metric to quantitatively compare the distribution of real anomaly scores with that of Gaussian-distributed pseudo-anomalies. We can see ~\cref{tab:tvd_results}, the resulting TVD value is about \textbf{0.08}, suggesting that the Gaussian-distributed pseudo-anomalies exhibit a significant overlap with the actual anomalies. This supports the assumptions that Gaussian noise can be used to preliminarily simulate anomalous behavior for anomaly detection, which is usually used by \textit{GLASS}\cite{chen2024unifiedanomalysynthesisstrategy}, \textit{SimpleNet}\cite{Liu_2023_CVPR}, \textit{RealNet}\cite{zhang2024realnetfeatureselectionnetwork},and \textit{DDPM}~\cite{ho2020denoising}. Extended Discussion of Gaussian Noise Validation can be seen in ~\cref{sec:extendedgaussian} in \textit{Supplementary Materials}. 
\section{Conclusion}

In this work, we reconceptualize overfitting as a controllable and transformative mechanism, using \textit{COAD} to enhance model capabilities beyond conventional boundaries. We introduce \textit{ARQ} to precisely regulate overfitting, as well as \textit{RADI}, which leverages \textit{$ARQ_{optimal}$} to provide a more versatile metric compared to \textit{AUROC-pixel}, facilitating both theoretical modeling and \textit{Dual Control Mechanism}. By repurposing overfitting as a generalizable module, our approach not only dismantles its demonization but also enables us to achieve SOTA results in both \textit{one-class} and \textit{multi-class} anomaly detection tasks. Furthermore, we provide a robust theoretical foundation for employing Gaussian noise as a preliminary pseudo-anomaly generator, extending the applicability of our \textit{COAD}.

\bibliographystyle{ACM-Reference-Format}
\bibliography{main}

\clearpage
\appendix


\twocolumn[
  \begin{center}
    \LARGE\bfseries Appendix
  \end{center}
]
\section{Dual Control Mechanism Design Concept}

Overfitting has traditionally been viewed as a problem that hampers generalization—where the model memorizes noise and specific details of the training data, leading to poor performance on unseen examples. However, in the context of anomaly detection, we posit that overfitting can be harnessed as a \textit{feature}, not a flaw. The key lies in the concept of \textit{controllable overfitting}—managing overfitting in a way that makes the model more sensitive to subtle differences between normal and anomalous data.

In practical terms, overfitting can amplify the sensitivity of the model to slight deviations, which is particularly useful in anomaly detection scenarios, where anomalies are often defined by subtle discrepancies from the norm. Our approach aims to leverage this characteristic to enhance detection capability while avoiding the negative consequences of overfitting, such as loss of generalization. This balancing act necessitates a carefully designed mechanism, which we call the \textbf{\textit{Dual Control Mechanism}} in ~\cref{eq:dual_control_overfit}.

The \textbf{Dual Control Mechanism} is comprised of two complementary components: \textit{$ARQ_{optimal}$} in ~\cref{eq:optimal_arq} and \textit{Gradient-Guided Overfitting Control} in ~\cref{eq:control_overfit}. These components work together to ensure that overfitting remains within beneficial bounds, ultimately enhancing anomaly detection without sacrificing the ability to generalize effectively.

\paragraph{Aberrance Retention Quotient}
The \textit{ARQ} quantifies the degree of overfitting by capturing the relationship between the model’s fit to the training data and its potential for retaining useful deviations that indicate anomalies. We define an optimal interval for ARQ, denoted as \textit{ARQ\(_{optimal}\)}, which represents the range where overfitting is beneficial—significantly enhancing sensitivity to anomalies without compromising generalization.

During training, \textit{ARQ} is continuously monitored to ensure it remains within this optimal interval. If \textit{ARQ} moves outside the defined range, corrective actions are taken to bring it back. The optimal interval allows the model to exploit overfitting in a targeted manner, enhancing the model’s sensitivity towards subtle anomalies while minimizing the risk of memorizing noise.

\paragraph{Gradient-Guided Overfitting Control}
Alongside \textit{ARQ\(_{optimal}\)}, the \textit{Gradient-Guided Overfitting Control} is employed to supervise how overfitting impacts anomaly detection performance directly. This component utilizes the \textit{Relative Anomaly Distribution Index (RADI)} to quantify the overlap between normal and anomalous score distributions. By keeping track of the gradient of \textit{RADI}, we can ensure that the anomaly detection performance is on an upward trajectory during overfitting.

Specifically, we require that the gradient of \textit{RADI} with respect to \textit{ARQ} is non-negative, which indicates that the model’s ability to distinguish anomalies from normal data is improving as overfitting increases. If the gradient condition is violated, indicating that overfitting is leading to a decline in detection performance, additional measures are triggered.

\paragraph{Freeze Command Signals for Progressive Layer Freezing}
In situations where the \textit{Dual Control Mechanism} fails—meaning both the \textit{ARQ} exceeds the optimal range and the \textit{Gradient-Guided Overfitting Control} shows deteriorating performance—we initiate \textit{Freeze Command Signals} in. These signals are the model’s way of mitigating excessive overfitting by progressively freezing network layers.

\textit{Freeze Command Signals} begin by selectively freezing the lower-level feature extraction layers, which are primarily responsible for capturing fundamental visual features such as edges, textures, and basic shapes. These foundational features are generally invariant across different images and therefore require less flexibility once they are well-trained. By freezing these lower-level layers, we stabilize the feature extraction process, reducing the risk of the model overfitting to noise and minute variations in the training set. As training progresses and overfitting persists beyond the optimal range, additional higher-level layers are frozen progressively in a staged manner. This gradual freezing strategy ensures that while the foundational layers remain stable, the more abstract, task-specific layers retain flexibility long enough to refine their understanding of anomalies.

By progressively freezing layers, the model maintains stable low-level feature extraction while allowing the higher-level layers to refine their understanding of anomalies. This approach ensures that the model retains a balance between maintaining its basic feature extraction capabilities and enhancing its discriminative power for anomaly detection.

The \textbf{Dual Control Mechanism} thus provides a comprehensive way to manage overfitting during training. By monitoring both \textit{ARQ} and the \textit{RADI} gradient, and employing \textit{Freeze Command Signals} when necessary, our framework transforms overfitting from a challenge into an asset—amplifying the model’s ability to detect anomalies while maintaining robustness and generalization.

\section{Main Functional Pseudocode}

\begin{algorithm}[H]
\caption{Training Stage with Gaussian Noise, Teacher-Student Reconstruction, and Overfitting Control}
\label{alg:training_stage_detailed}
\begin{algorithmic}[1]
\REQUIRE Dataset $\mathcal{D}$, Initial Model $M$, Learning Rate $\alpha$, Optimal ARQ Range $\theta_{optimal}$, Noise Standard Deviation $\sigma_{noise}$, Epochs $N_{standard}$, Overfit Epochs $N_{overfit}$
\STATE \textbf{Step 1: Standard Training Stage}
\FOR{epoch = 1 to $N_{standard}$}
    \FOR{batch in $\mathcal{D}_{train}$}
        \STATE Generate pseudo-anomalous data $\mathcal{D}_{anomaly}$ using Gaussian noise.
        \STATE Generate feature maps $F_{T}$ using the Teacher Model $M_{T}$ for training samples.
        \STATE Generate reconstructed feature maps $F_{S}$ using the Student Model $M_{S}$.
        \STATE Compute $L_{reconstruction}$ between $F_{T}$ and $F_{S}$ and other losses.
        \STATE Update Models
    \ENDFOR
\ENDFOR

\STATE \textbf{Step 2: Controllable Overfitting Stage}
\STATE Set learning rate $\alpha_{overfit} = \alpha / 10$
\STATE Initialize Freeze Counter $C = 0$
\FOR{epoch = 1 to $N_{overfit}$}
    \FOR{batch in $\mathcal{D}_{train}$}
        \STATE Generate pseudo-anomalous data $\mathcal{D}_{anomaly}$ using Gaussian noise.
        \STATE Generate feature maps $F_{T}$ using the Teacher Model $M_{T}$ for both normal and pseudo-anomalous samples.
        \STATE Generate reconstructed feature maps $F_{S}$ using the Student Model $M_{S}$.
        \STATE Calculate ARQ 
        \STATE Compute RADI based on score distributions

        \STATE \textbf{Dual Control Mechanism:}
        \IF{$\theta$ exceeds $\theta_{optimal}$ or $\frac{dRADI(\theta)}{d\theta} < 0$}
            \STATE Increment Freeze Counter $C$
            \IF{$C > C_thr$}
                \STATE Issue \textit{Freeze Command Signals}
                \STATE \textbf{Selective Freezing}
            \ENDIF
        \ELSE
            \STATE Reset Freeze Counter $C$
        \ENDIF

        \STATE Compute $L_{reconstruction}$ between $F_{T}$ and $F_{S}$ and other losses.
        
        \STATE Update Models
    \ENDFOR
\ENDFOR
\end{algorithmic}
\end{algorithm}

\begin{algorithm}[H]
\caption{Selective Freezing}
\label{alg:selective_freezing}
\begin{algorithmic}[1]
\REQUIRE Student Model $M_{S}$, Freeze Counter $C$, Threshold $C_{max}$, Layer List $L$, ARQ $\theta$, Optimal ARQ Range $\theta_{optimal}$
\STATE Initialize $C = 0$ \COMMENT{Freeze Counter initialized to zero}
\FOR{epoch = 1 to $N_{overfit}$}
    \FOR{batch in training data}
        \STATE Perform forward pass and compute ARQ $\theta$
        
        \IF{$\theta$ exceeds $\theta_{optimal}$ \textbf{or} $\frac{dRADI(\theta)}{d\theta} < 0$}
            \STATE Increment Freeze Counter: $C = C + 1$
        \ELSE
            \STATE Reset Freeze Counter: $C = 0$
        \ENDIF

        \IF{$C > C_{thr}$}
            \STATE Issue \textit{Freeze Command Signal} to initiate layer freezing
            \STATE \textbf{Layer Freezing Process:}
                \FOR{layer $l$ in $L$ (ordered from lowest-level to highest-level)}
                    \IF{layer $l$ is not frozen}
                        \STATE \textbf{Freeze Procedure for Current Layer}:
                        \STATE $\quad \forall p \in \text{parameters of layer $l$}$
                        \STATE $\quad \quad p.requires\_grad = \textbf{False}$ \COMMENT{Stop gradient updates for layer $l$}
                        
                        \STATE Update training status: mark layer $l$ as \textit{frozen}
                        
                        \STATE \textbf{Apply Forward Freezing Effects:}
                        \STATE $\quad$ Notify subsequent layers in network to adjust activations \COMMENT{Maintain balance during backpropagation}
                        \STATE $\quad$ Update batch normalization layers if necessary \COMMENT{Handle training vs. inference mode}
                        
                        \STATE Reset Freeze Counter: $C = 0$ \COMMENT{Reset counter after successfully freezing a layer}
                        
                        \STATE Add layer $l$ to the Frozen Layer List $F$ \COMMENT{Keep track of all frozen layers}
                        
                        \STATE \textbf{Break} from the loop \COMMENT{Freeze only one layer in each iteration to avoid rapid performance drops}
                    \ELSE
                        \STATE \textbf{Continue} to next layer \COMMENT{Skip if layer is already frozen}
                    \ENDIF
                \ENDFOR
        \ENDIF

        \STATE \textbf{Continue Training:}
        \STATE Perform backpropagation and update weights accordingly
    \ENDFOR
\ENDFOR
\end{algorithmic}
\end{algorithm}

\begin{algorithm}[H]
\caption{Inference Stage}
\label{alg:inference_phase}
\begin{algorithmic}[1]
\REQUIRE Trained Teacher Model $M_{\text{teacher}}$, Trained Student Model $M_{\text{student}}$, Test Dataset $\mathcal{D}_{test}$, Optional Pseudo Anomaly Filter $F$
\STATE \textbf{Initialization:} Set models in evaluation mode.
\FOR{test sample $x$ in $\mathcal{D}_{test}$}
    \STATE \textbf{Input Feature Extraction:}
    \STATE \quad Extract features using teacher model: $f_{\text{teacher}} \leftarrow M_{\text{teacher}}(x)$
    
    \STATE \textbf{Student Model Inference:}
    \STATE \quad Extract reconstructed features using student model: $\hat{f}_{\text{student}} \leftarrow M_{\text{student}}(f_{\text{filtered}})$
    
    \STATE \textbf{Reconstruction Computation:}
    \STATE \quad Compute reconstruction error to classify sample: $e(x) = \text{Loss}(x, \hat{f}_{\text{student}})$
    
    \STATE \textbf{Prediction Generation:}
    \IF{$e(x) > \text{Threshold}$}
        \STATE Classify as \textbf{Anomalous}
    \ELSE
        \STATE Classify as \textbf{Normal}
    \ENDIF
    
    \STATE \textbf{Store Results:} Save prediction results for sample $x$.
\ENDFOR
\STATE \textbf{Output:} Anomaly classification for all test samples in $\mathcal{D}_{test}$
\end{algorithmic}
\end{algorithm}

\begin{algorithm}[H]
\caption{Calculation of Aberrance Retention Quotient (\textit{ARQ})}
\label{alg:arq_calculation}
\begin{algorithmic}[1]
\REQUIRE Predicted values $\hat{y}_i$, Ground truth values $y_i$, Total instances $N$
\STATE Initialize \textit{ARQ} value: $\textit{ARQ} \gets 0$
\STATE \textbf{Compute Numerator:}
\STATE \quad $Numerator \gets \sum_{i=1}^{N} |\hat{y}_i - y_i|$
\STATE \textbf{Compute Denominator:}
\STATE \quad $Denominator \gets \sum_{i=1}^{N} y_i$
\STATE \textbf{Calculate ARQ:}
\STATE \quad $\textit{ARQ} \gets \frac{Numerator}{Denominator}$
\STATE \textbf{Output:} Final \textit{ARQ} value
\end{algorithmic}
\end{algorithm}

\begin{algorithm}[H]
\caption{Calculation of Relative Anomaly Distribution Index (\textit{RADI})}
\label{alg:radi_calculation}
\begin{algorithmic}[1]
\REQUIRE Scores of anomalous pixels $S_a$, Scores of normal pixels $S_n$
\STATE Initialize \textit{RADI} value: $\textit{RADI} \gets 0$
\STATE \textbf{Compute \textit{RADI} using Integral:}
\STATE \quad $\textit{RADI} \gets \int_{-\infty}^{\infty} P(S_a > x) \cdot f_{S_n}(x) \, dx$
\STATE \textbf{Output:} Final \textit{RADI} value
\end{algorithmic}
\end{algorithm}

\section{Extended Discussion of Gaussian Noise Validation}
\label{sec:extendedgaussian}
\subsection{Theoretical Rationale for Gaussian Noise}
In this work, we choose to use Gaussian noise as a preliminarily pseudo-anomaly generator due to its statistical properties and simplicity. Gaussian noise, characterized by its mean and variance, serves as an ideal candidate for generating random deviations that can mimic unexpected patterns in normal data. 

From a theoretical standpoint, Gaussian noise exhibits a distribution that is both isotropic and well-defined in high-dimensional space, making it suitable for approximating unstructured, irregular anomalies. The Central Limit Theorem (CLT) supports the use of Gaussian distributions as approximations for many naturally occurring random processes. In particular, when adding minor perturbations to an image, we aim to simulate unforeseen deviations from the underlying structure of normal samples, which Gaussian noise effectively captures.

Additionally, Gaussian noise serves as a zero-mean stochastic process, allowing us to avoid introducing any specific directional bias, which could potentially interfere with the model's capacity to learn anomalous versus normal regions. By employing Gaussian noise, we ensure that our model does not overfit to particular anomaly patterns but instead gains a generalized sensitivity to any divergence from the normal data distribution. This universality underpins our approach to controllable overfitting within the \textit{COAD} framework.

\subsection{Validation Through Total Variation Distance (TVD)}
To validate the suitability of Gaussian noise as a pseudo-anomaly generator, we conduct an empirical analysis using the \textit{Total Variation Distance (TVD)} metric. TVD is a standard measure used in probability theory to quantify the dissimilarity between two distributions. We calculate TVD to evaluate how closely Gaussian noise could mimic true anomalies in both spatial and intensity domains.

The calculation of TVD involves estimating the difference between the empirical distributions of Gaussian noise and the true normal and anomaly pixel intensities:

\begin{equation}
    D_{\text{TVD}}(P, Q) = \frac{1}{2} \sum_x |P(x) - Q(x)|,
\end{equation}

\noindent where $P$ represents the empirical distribution of the true anomalies, and $Q$ represents the empirical distribution of Gaussian noise. TVD measures the extent of overlap between the two distributions, where a smaller TVD implies a closer approximation between the pseudo-anomalies and real anomalies.

We conduct experiments on our dataset using both generated Gaussian noise and real anomalies.  presents a comparison of the TVD values for different classes. The results indicate that the Gaussian distribution closely matches the distribution of real normal and anomalies across a range of classes. Specifically, it can be seen in ~\cref{sec:val_gau}.

\begin{table*}[htbp]
\centering
\caption{Performance comparison across different methods in \textit{multi-class} tasks on \textit{VisA} dataset.}
\label{tab:multi-class-visa}
\resizebox{\textwidth}{!}{
\begin{tabular}{c|c|ccc|cc}
\toprule
 \multicolumn{2}{c|}{\textit{Category}} & \textit{DRAEM}~\cite{Zavrtanik_2021_ICCV}& \textit{JNLD}~\cite{Liu_2023_CVPR}  & \textit{OmniAL}~\cite{Zhao_2023_CVPR} &\textit{UniAD}~\cite{you2022unifiedmodelmulticlassanomaly}
&\textit{UniAD}(+\textit{COAD})

\\ 
\midrule
\multirow{4}{*}{\textit{Single Instance}}
 &\textit{Cashew}       
& 67.3 / 68.5  & 82.5 / 94.1  & 88.6 / 95.0  & 92.8/\underline{\textbf{98.6}} 
& \underline{\textbf{93.7}}/\underline{\textbf{98.6}}            
\\

 &\textit{Chewing gum}  
& 90.0 / 92.7  & 96.0 / 98.9  & 96.4 / \textbf{99.0}  & 96.3/98.8           
& \underline{\textbf{96.6}}/\underline{\textbf{99.0}}         
\\

 &\textit{Fryum}        
& 86.2 / 83.2  & 91.9 / 90.0  & \textbf{94.6} / 92.1  & 83.0/95.9          
& \underline{87.9}/\underline{\textbf{97.3}}         
\\
 &\textit{Pipe fryum}   
& 87.1 / 72.3  & 87.5 / 92.5  & 86.1 / 98.2  & 94.7/\underline{\textbf{98.9}}          
& \underline{\textbf{97.8}}/98.4          
\\
\midrule
\multirow{4}{*}{\textit{Multiple Instances}}
 &\textit{Candle}       
& 70.2 / 82.6  & 85.4 / 94.5  & 86.8 / 95.8  & 94.1/98.5           
& \underline{\textbf{95.8}}/\underline{\textbf{98.7}}           
\\

 &\textit{Capsules}     
& 89.6 / 96.6 & \textbf{91.4} / \textbf{99.6}  & 90.6 / 99.4  & 55.6/88.7          
& \underline{78.3}/\underline{97.1}         
\\
 &\textit{Macaroni 1}   
& 68.6 / 89.8  & 90.5 / 93.3  & \textbf{92.6} / \textbf{98.6}  & 79.9/97.4          
& \underline{87.2}/\underline{98.2}         
\\

 &\textit{Macaroni 2}   
& 60.3 / 83.2  & 71.3 / 92.1  & 75.2 / \textbf{97.9}  & 71.6/95.2          
& \underline{\textbf{80.4}}/\underline{97.6}         
\\
\midrule
\multirow{4}{*}{\textit{Complex Structure}}
 &\textit{PCB1}         
& 83.9 / 94.0  & 82.9 / \textbf{98.0}  & 77.7 / 97.6  & 92.8/93.3          
& \underline{\textbf{96.9}}/\underline{97.1}         
\\

 &\textit{PCB2}         
& 81.7 / 94.1  & 79.1 / 95.0  & 81.0 / 93.9  & 87.8/93.9          
& \underline{\textbf{94.1}}/\underline{\textbf{95.2}}         
\\

 &\textit{PCB3}         
& 87.7 / 94.1  & \textbf{90.1} / \textbf{98.5}  & 88.1 / 94.7  & 78.6/97.3          
& \underline{84.5}/\underline{98.0}         
\\

 &\textit{PCB4}         
& 87.1 / 72.3  & 96.2 / \textbf{97.5}  & 95.3 / 97.1  & \underline{\textbf{98.8}}/\underline{94.9}          
& 98.7/94.7          
\\
\midrule
\rowcolor[gray]{0.9}
\multicolumn{2}{c|}{\textit{\textbf{Avg.}}}           & 80.0/85.3                   & 87.1/95.3& 87.8/96.6& 85.5/95.9          &\underline{91.0}/\underline{97.5}         \\ 
\bottomrule
\end{tabular}}
\end{table*}

\subsection{Experimental Results and Qualitative Study}
The experimental results demonstrate that the model trained using Gaussian noise exhibits a better anomaly detection performance to models trained with other pseudo-anomaly generator.

\begin{figure*}[htbp]
    \centering
    % 设置子图大小和行间距
    \setlength{\tabcolsep}{4pt} % 子图列间距
    \renewcommand{\arraystretch}{1.5} % 调整行间距

    % Anomaly 子图部分
    \begin{tabular}{ccc}
        \subcaptionbox{Bottle (Anomaly)}{
            \includegraphics[width=0.3\textwidth]{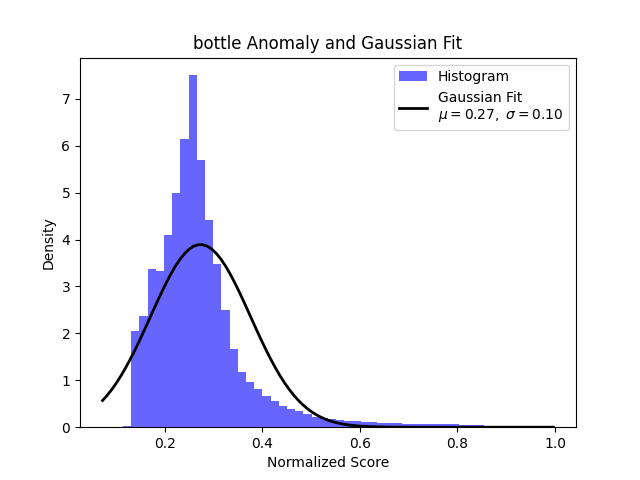}
        } &
        \subcaptionbox{Cable (Anomaly)}{
            \includegraphics[width=0.3\textwidth]{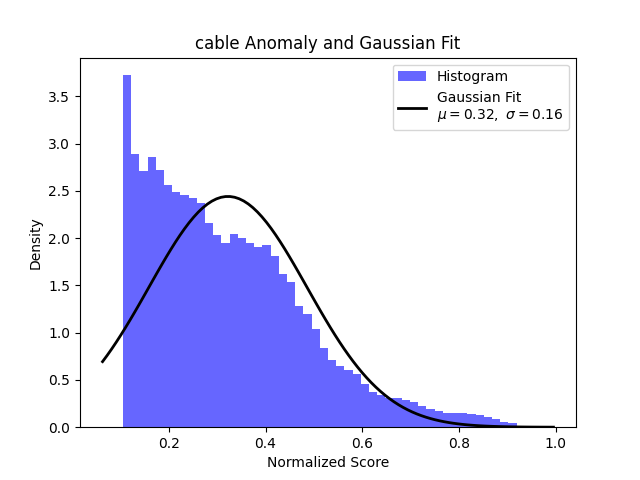}
        } &
        \subcaptionbox{Capsule (Anomaly)}{
            \includegraphics[width=0.3\textwidth]{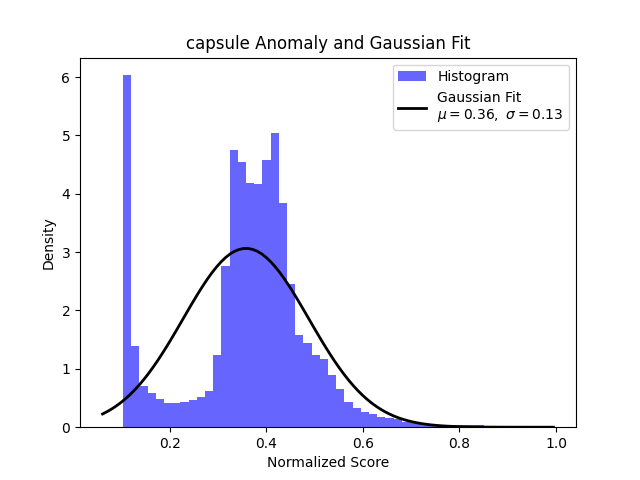}
        } \\
        \subcaptionbox{Carpet (Anomaly)}{
            \includegraphics[width=0.3\textwidth]{figure/carpet_anomaly.png}
        } &
        \subcaptionbox{Grid (Anomaly)}{
            \includegraphics[width=0.3\textwidth]{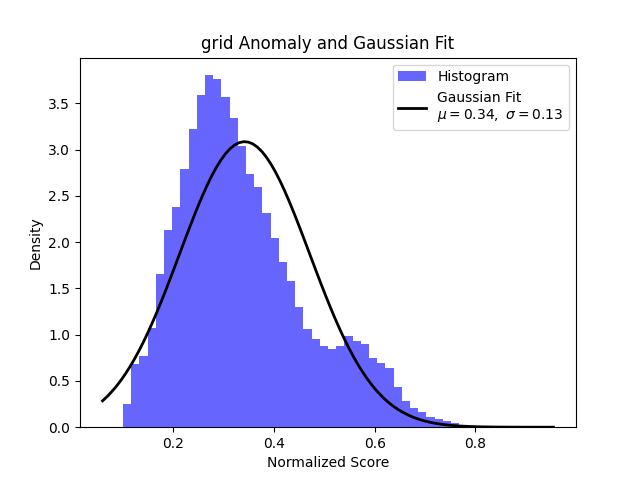}
        } &
        \subcaptionbox{Hazelnut (Anomaly)}{
            \includegraphics[width=0.3\textwidth]{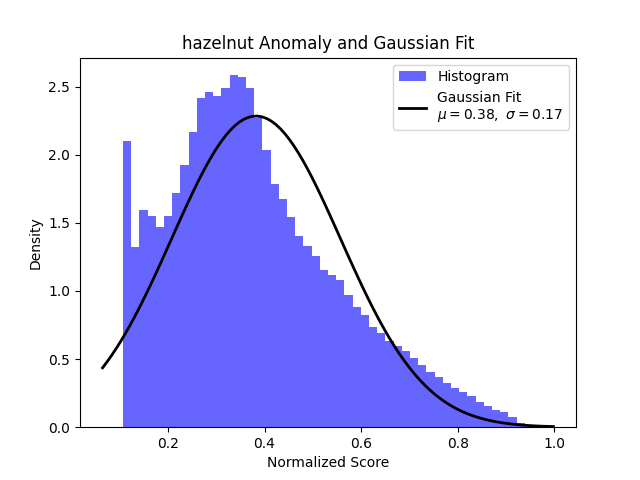}
        } \\
        \subcaptionbox{Leather (Anomaly)}{
            \includegraphics[width=0.3\textwidth]{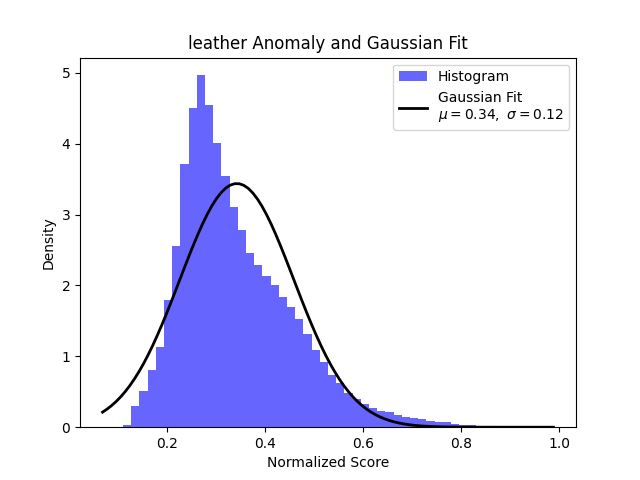}
        } &
        \subcaptionbox{Metal Nut (Anomaly)}{
            \includegraphics[width=0.3\textwidth]{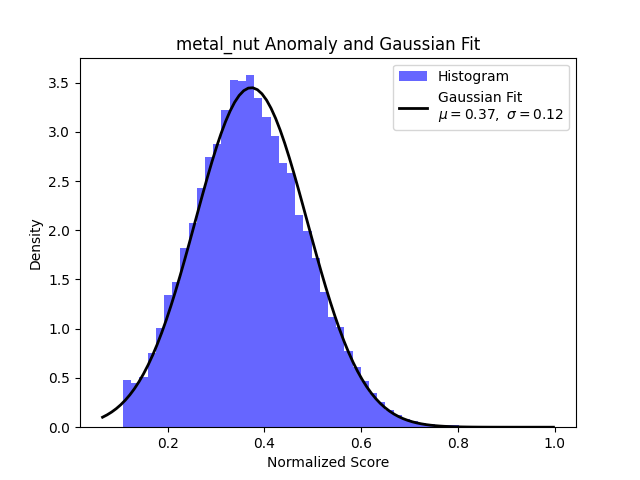}
        } &
        \subcaptionbox{Pill (Anomaly)}{
            \includegraphics[width=0.3\textwidth]{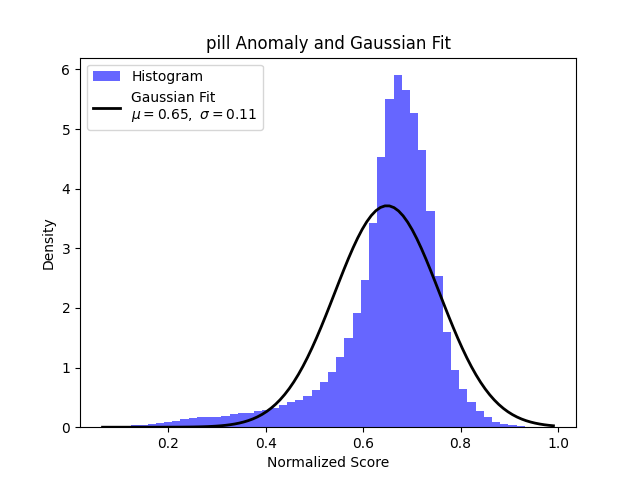}
        } \\
        \subcaptionbox{Screw (Anomaly)}{
            \includegraphics[width=0.3\textwidth]{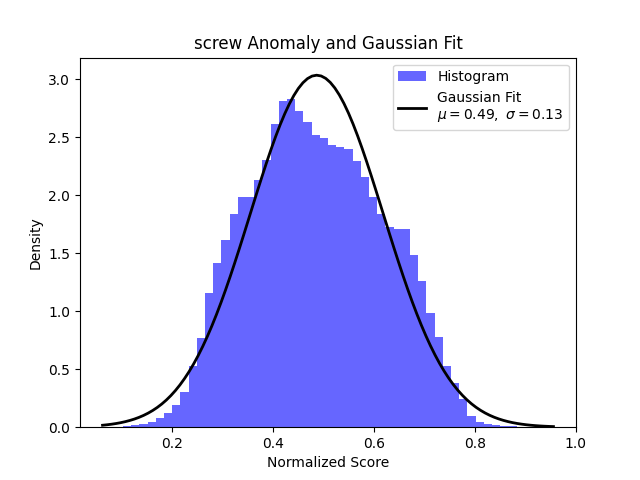}
        } &
        \subcaptionbox{Tile (Anomaly)}{
            \includegraphics[width=0.3\textwidth]{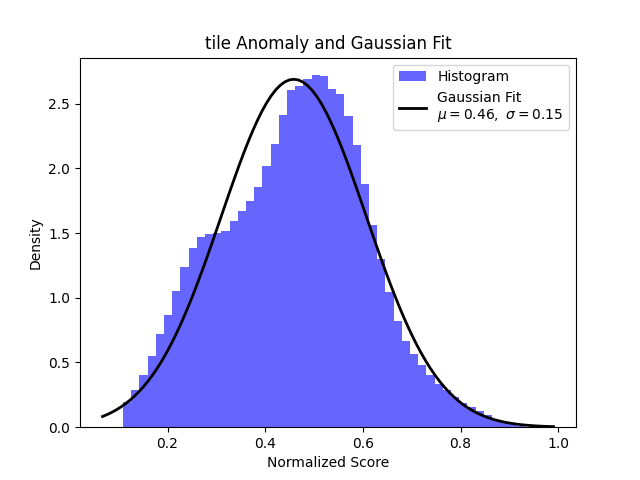}
        } &
        \subcaptionbox{Toothbrush (Anomaly)}{
            \includegraphics[width=0.3\textwidth]{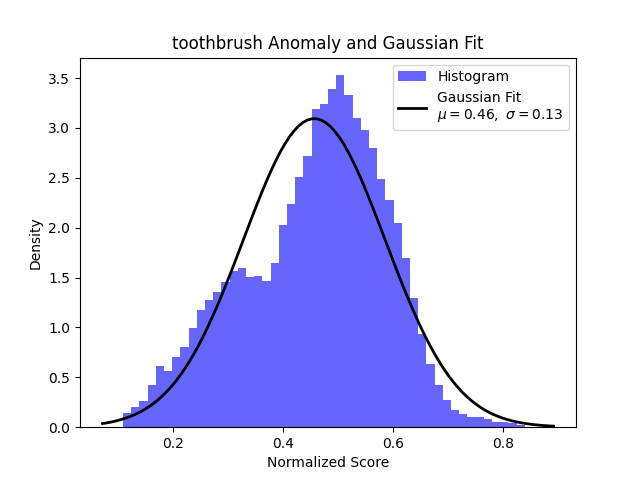}
        } \\
        \subcaptionbox{Transistor (Anomaly)}{
            \includegraphics[width=0.3\textwidth]{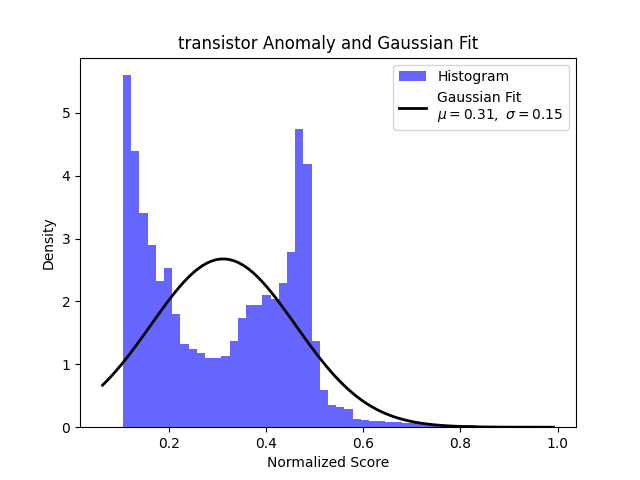}
        } &
        \subcaptionbox{Wood (Anomaly)}{
            \includegraphics[width=0.3\textwidth]{figure/wood_anomaly.png}
        } &
        \subcaptionbox{Zipper (Anomaly)}{
            \includegraphics[width=0.3\textwidth]{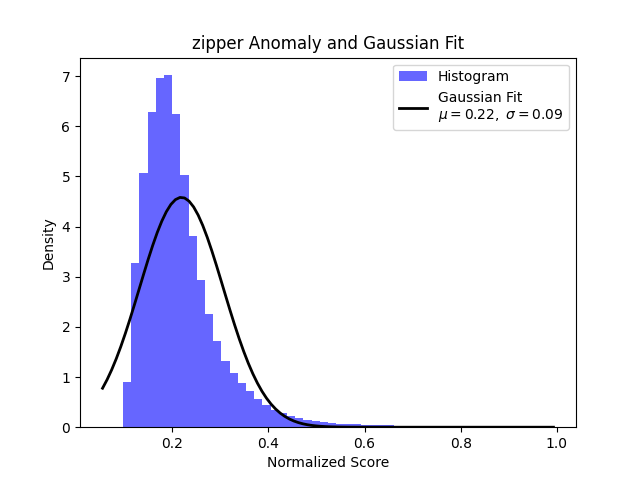}
        } \\
    \end{tabular}
    \caption{Histograms for anomaly scores of 15 individual categories. Each histogram includes Gaussian fitting curves for better visualization.}
    \label{fig:anomaly_histograms}
\end{figure*}

\begin{figure*}[htbp]
    \centering
    % 设置子图大小和行间距
    \setlength{\tabcolsep}{4pt} % 子图列间距
    \renewcommand{\arraystretch}{1.5} % 调整行间距

    % Normal 子图部分
    \begin{tabular}{ccc}
        \subcaptionbox{Bottle (Normal)}{
            \includegraphics[width=0.3\textwidth]{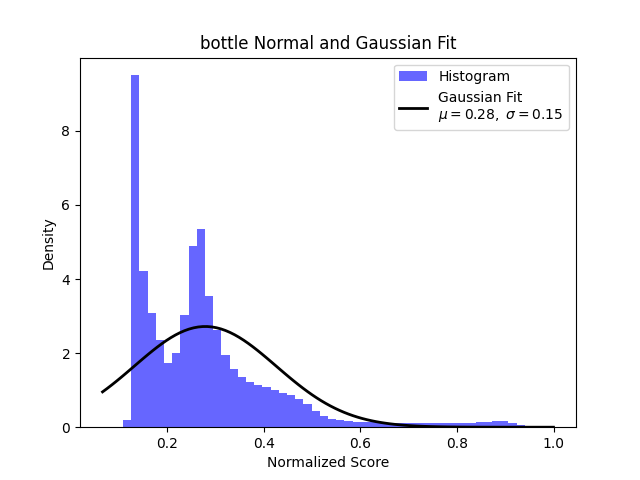}
        } &
        \subcaptionbox{Cable (Normal)}{
            \includegraphics[width=0.3\textwidth]{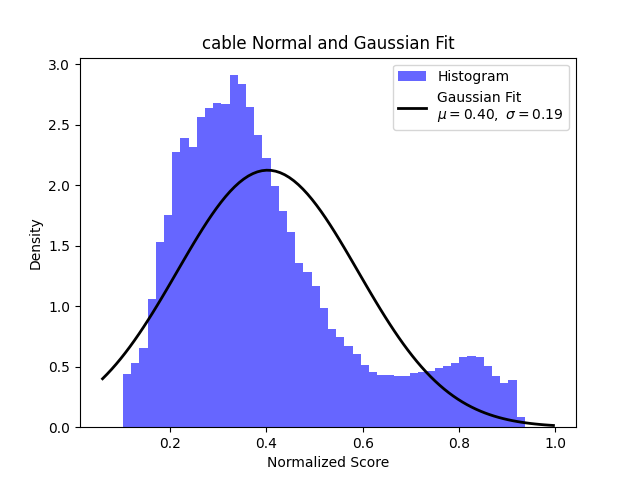}
        } &
        \subcaptionbox{Capsule (Normal)}{
            \includegraphics[width=0.3\textwidth]{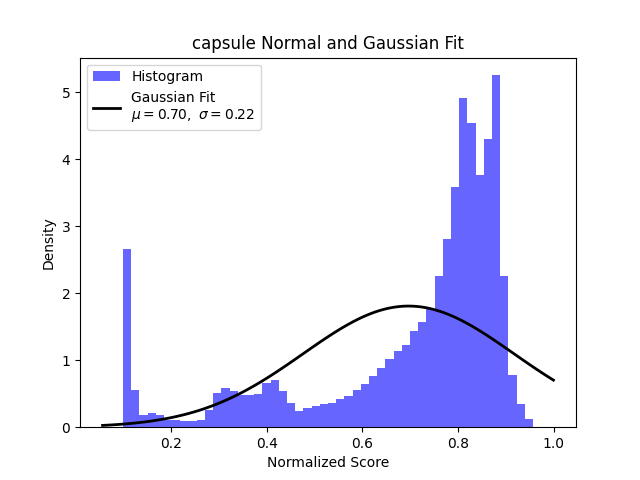}
        } \\
        \subcaptionbox{Carpet (Normal)}{
            \includegraphics[width=0.3\textwidth]{figure/carpet_normal.png}
        } &
        \subcaptionbox{Grid (Normal)}{
            \includegraphics[width=0.3\textwidth]{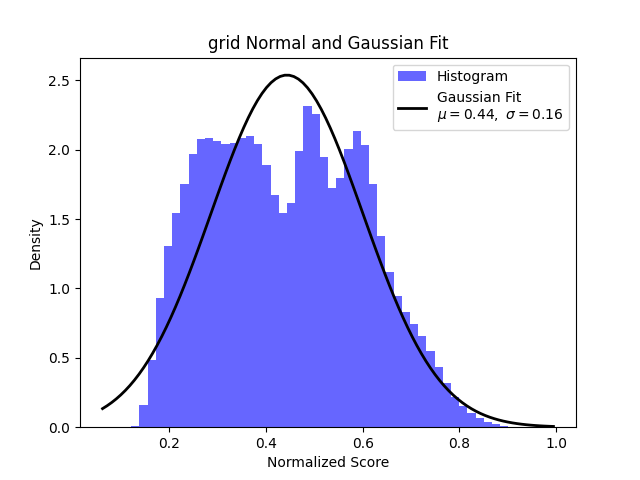}
        } &
        \subcaptionbox{Hazelnut (Normal)}{
            \includegraphics[width=0.3\textwidth]{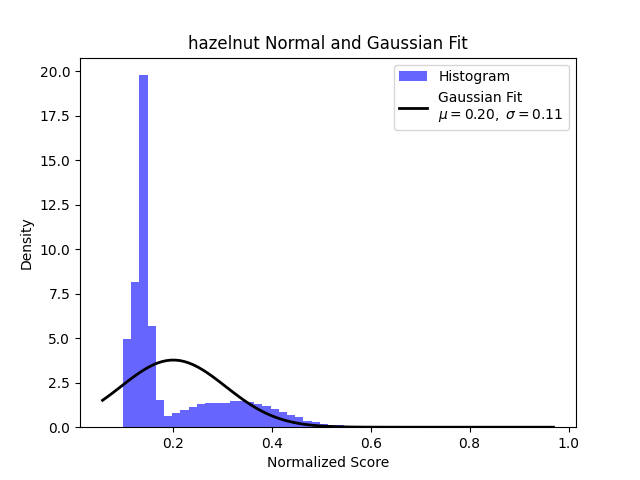}
        } \\
        \subcaptionbox{Leather (Normal)}{
            \includegraphics[width=0.3\textwidth]{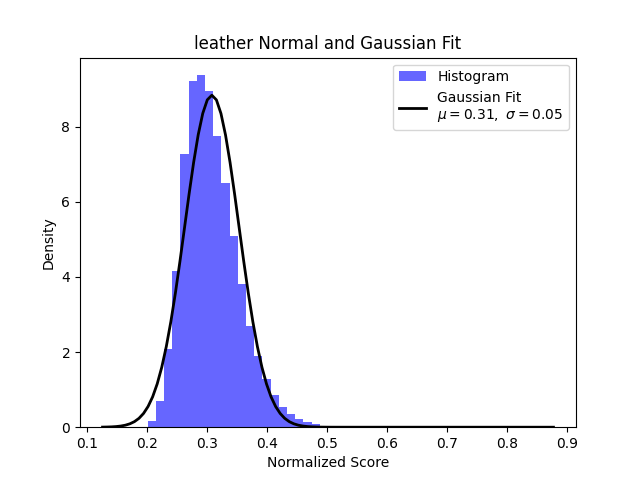}
        } &
        \subcaptionbox{Metal Nut (Normal)}{
            \includegraphics[width=0.3\textwidth]{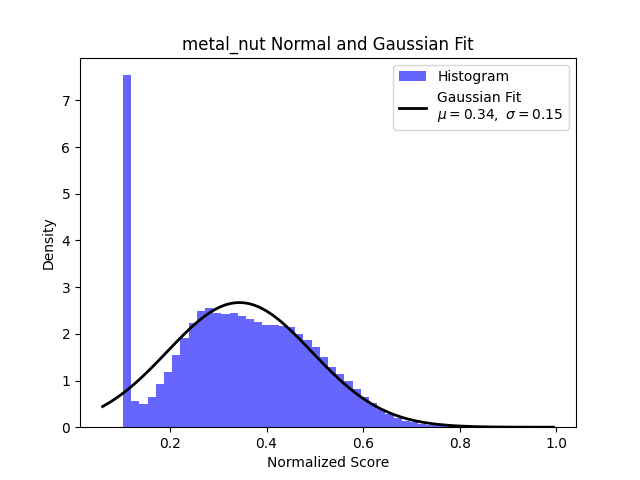}
        } &
        \subcaptionbox{Pill (Normal)}{
            \includegraphics[width=0.3\textwidth]{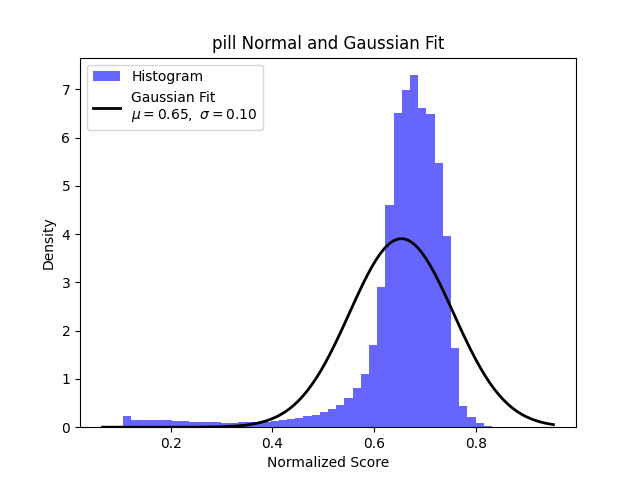}
        } \\
        \subcaptionbox{Screw (Normal)}{
            \includegraphics[width=0.3\textwidth]{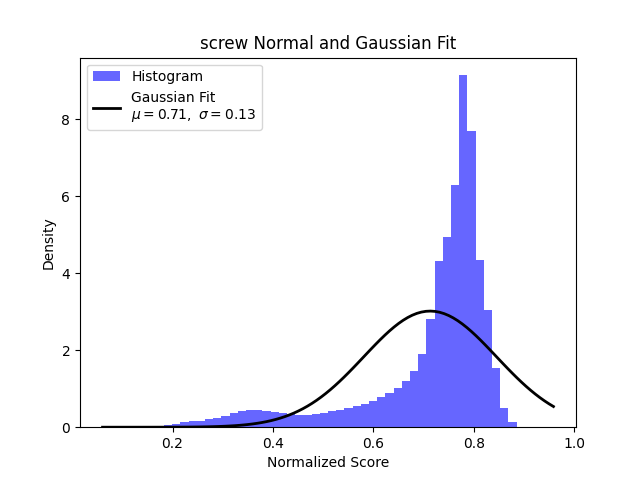}
        } &
        \subcaptionbox{Tile (Normal)}{
            \includegraphics[width=0.3\textwidth]{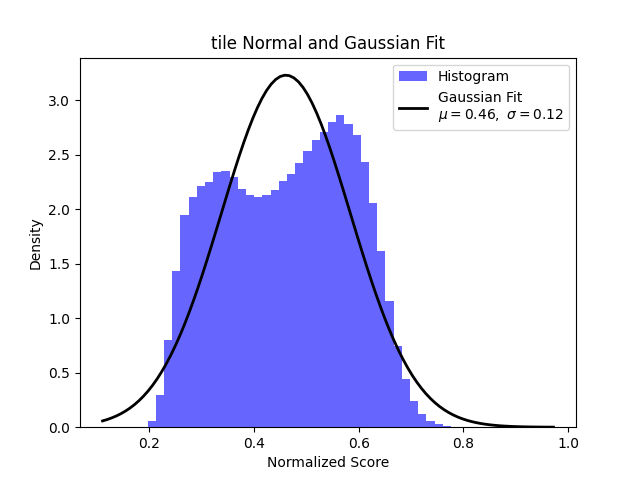}
        } &
        \subcaptionbox{Toothbrush (Normal)}{
            \includegraphics[width=0.3\textwidth]{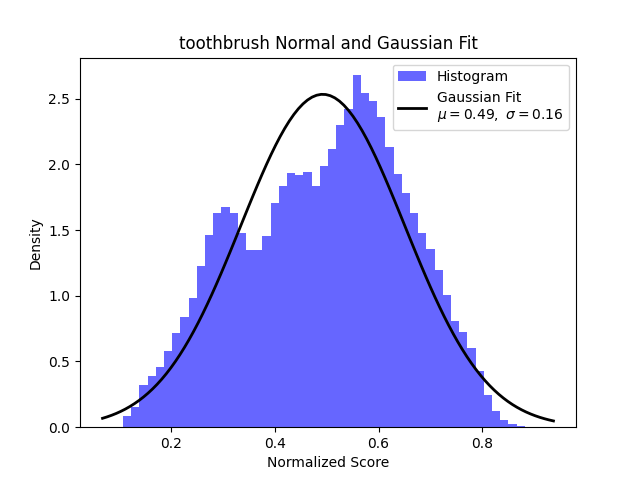}
        } \\
        \subcaptionbox{Transistor (Normal)}{
            \includegraphics[width=0.3\textwidth]{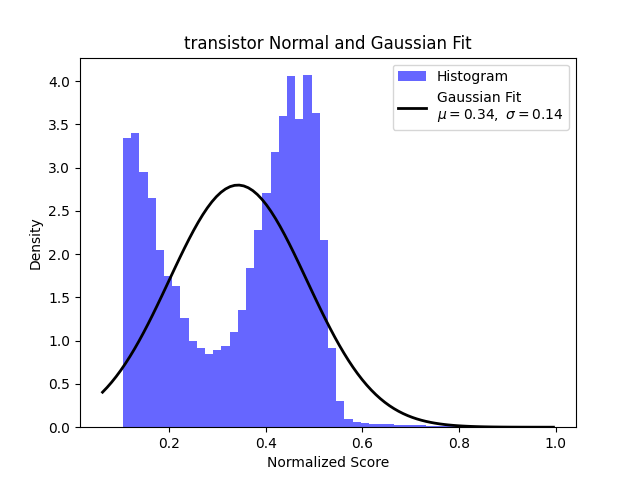}
        } &
        \subcaptionbox{Wood (Normal)}{
            \includegraphics[width=0.3\textwidth]{figure/wood_normal.png}
        } &
        \subcaptionbox{Zipper (Normal)}{
            \includegraphics[width=0.3\textwidth]{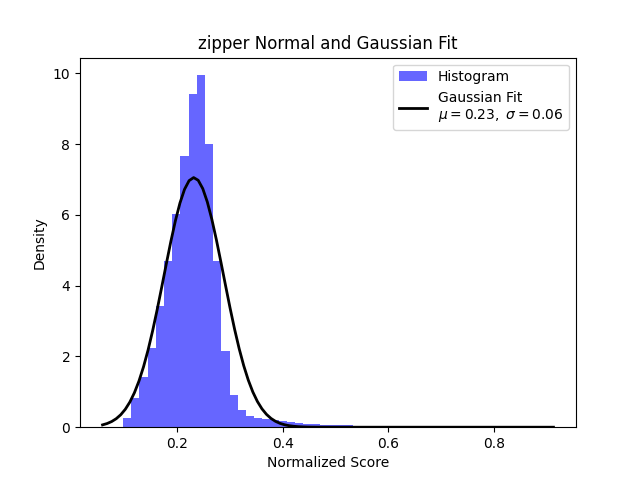}
        } \\
    \end{tabular}
    \caption{Histograms for normal scores of 15 individual categories. Each histogram includes Gaussian fitting curves for better visualization.}
    \label{fig:normal_histograms}
\end{figure*}

To further validate the results, we present detailed histograms for the anomaly detection scores obtained across 15 individual categories as well as the overall dataset. These histograms are generated under the condition where the \textit{Anomaly Rate Quotient (ARQ)} is set to 0.006, using \textit{RD++} as our baseline framework. Each histogram in ~\cref{fig:anomaly_histograms,fig:normal_histograms} provides a clear visualization of the distribution of detection scores, overlaid with a Gaussian fitting curve for enhanced interpretability. This fitting curve highlights the distribution, allowing for a more effective comparison of the model's performance across different categories and the overall dataset, further solidifying the reliability of our evaluation metrics.

\subsection{Future Directions with GLASS-like Pseudo-Anomaly Generators}
\label{sec:glass-like}
While Gaussian noise has proven effective as a preliminary pseudo-anomaly generator, future work could explore more advanced methods inspired by the \textit{GLASS} approach~\cite{chen2024unifiedanomalysynthesisstrategy}. The \textit{GLASS} method introduces a structured way to generate pseudo-anomalies by leveraging geometric transformations and localized pattern synthesis, providing greater control over anomaly characteristics.

By incorporating concepts from \textit{GLASS}, we envision a \textit{Gaussian Noise-Enhanced GLASS-like Generator}, where Gaussian noise could act as the foundational perturbation, and additional constraints or transformations could be applied to shape the pseudo-anomalies in a contextually relevant manner. For instance, augmenting Gaussian noise with spatial or intensity correlations reflective of real anomalies could enhance its realism and improve model robustness in detecting diverse anomalies.

Such an approach would align well with our \textit{COAD} framework, enabling a more versatile training process that accounts for both unstructured and structured anomaly patterns. By systematically integrating these future developments, we aim to refine our anomaly detection strategy, bridging the gap between theoretical advancements and practical applications.

\section{Additional Experimental Results on VisA Dataset} 
\label{sec:additional_visa}

In this section, we provide additional quantitative results for the \textit{multi-class anomaly detection} task evaluated on the VisA dataset, which complements the results presented in the main paper. Detailed performance comparisons across different methods are summarized in Table~\ref{tab:multi-class-visa}.

\section{Qualitative Comparison of Anomaly Detection between Different Baselines and COAD}
\label{sec:qualitativecomparison}

\begin{figure*}[htbp]
    \centering
    % 设置子图大小和行间距
    \setlength{\tabcolsep}{4pt} % 子图列间距
    \renewcommand{\arraystretch}{1.5} % 调整行间距

    \begin{tabular}{cc} % 两列布局
        \subcaptionbox{Bottle}{
            \includegraphics[width=0.45\textwidth]{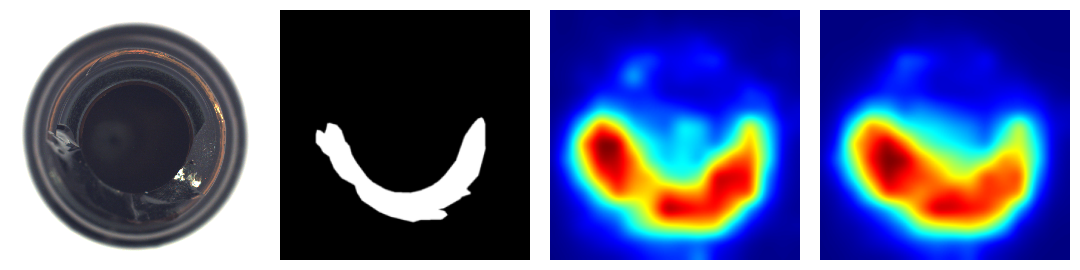}
        } &
        \subcaptionbox{Cable}{
            \includegraphics[width=0.45\textwidth]{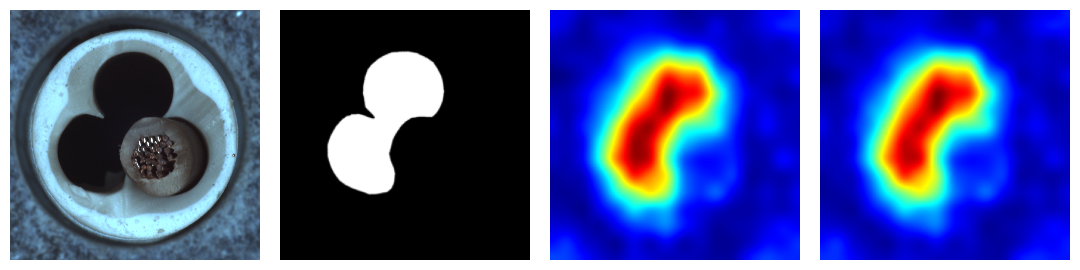}
        } \\
        \subcaptionbox{Capsule}{
            \includegraphics[width=0.45\textwidth]{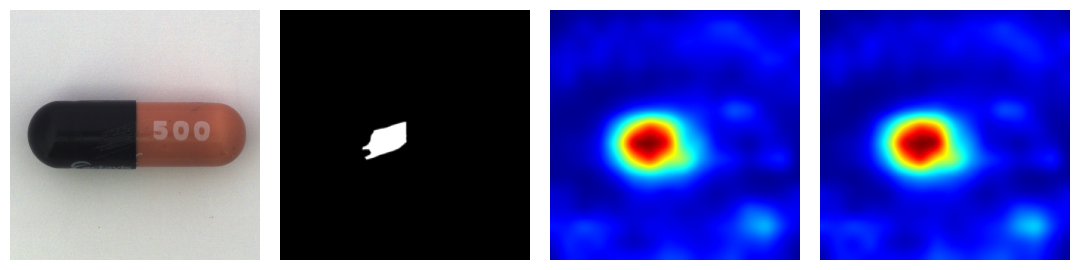}
        } &
        \subcaptionbox{Carpet\label{fig:carpet}}{
            \includegraphics[width=0.45\textwidth]{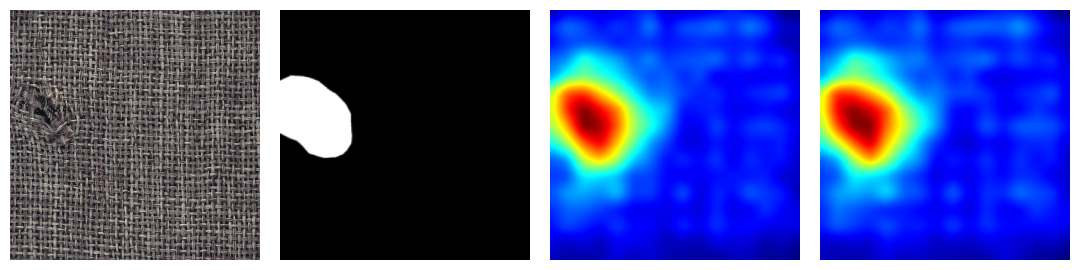}
        } \\
        \subcaptionbox{Grid\label{fig:grid}}{
            \includegraphics[width=0.45\textwidth]{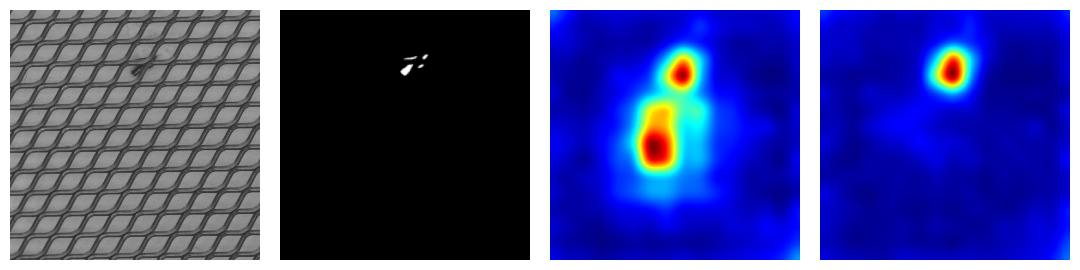}
        } &
        \subcaptionbox{Hazelnut\label{fig:hazelnut}}{
            \includegraphics[width=0.45\textwidth]{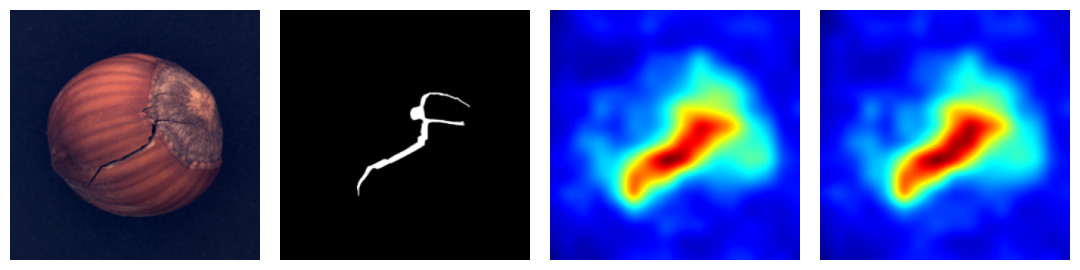}
        } \\
        \subcaptionbox{Leather\label{fig:leather}}{
            \includegraphics[width=0.45\textwidth]{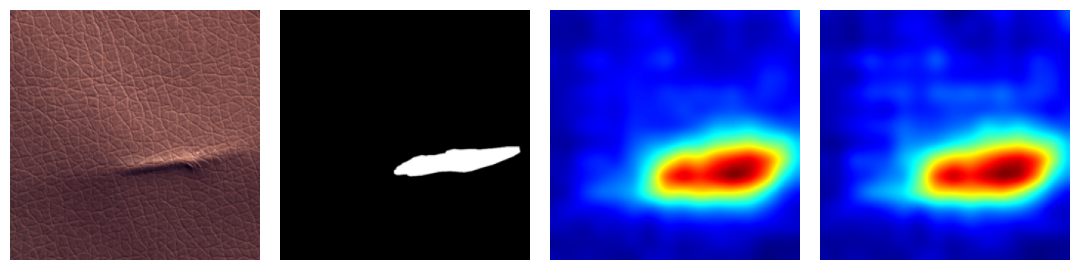}
        } &
        \subcaptionbox{Metalnut\label{fig:metal_nut}}{
            \includegraphics[width=0.45\textwidth]{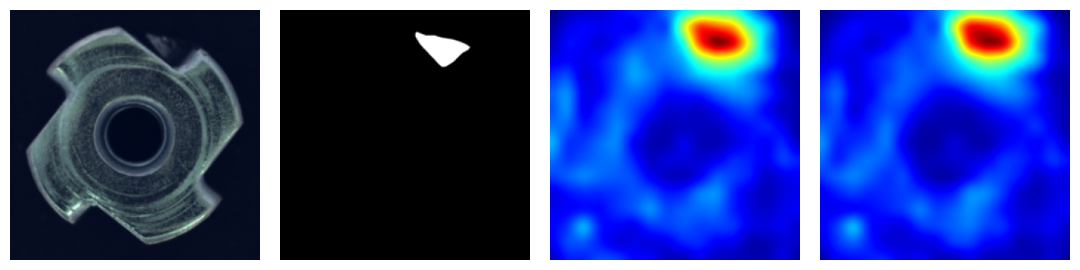}
        } \\
        \subcaptionbox{Pill\label{fig:pill}}{
            \includegraphics[width=0.45\textwidth]{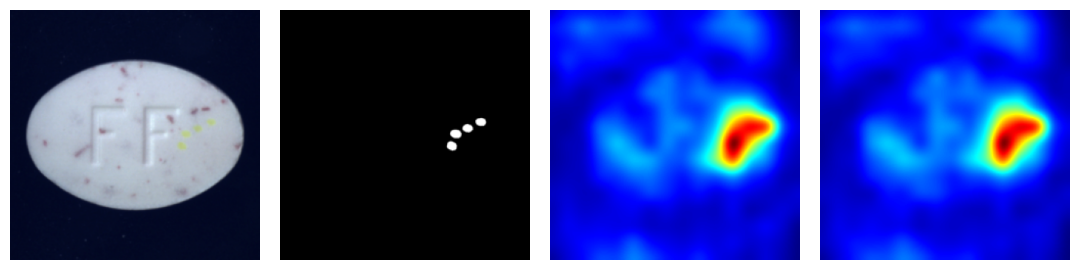}
        } &
        \subcaptionbox{Screw\label{fig:screw}}{
            \includegraphics[width=0.45\textwidth]{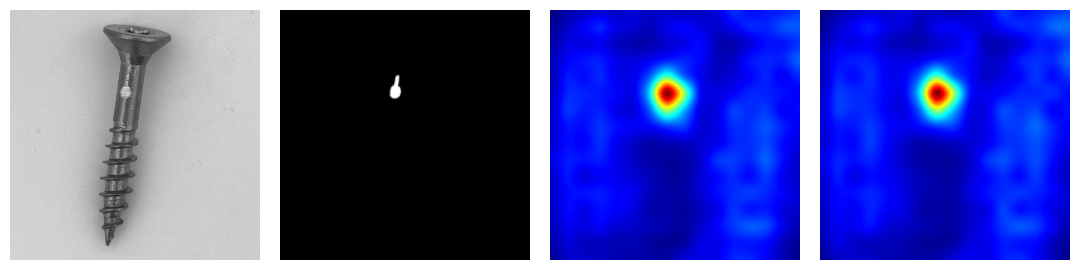}
        } \\
        \subcaptionbox{Tile\label{fig:tile}}{
            \includegraphics[width=0.45\textwidth]{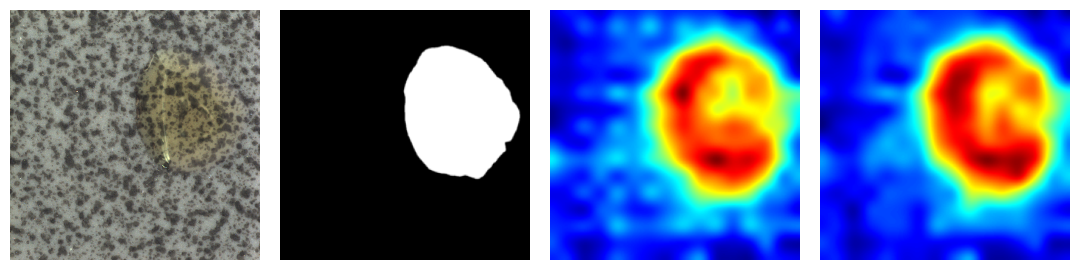}
        } &
        \subcaptionbox{Toothbrush\label{fig:toothbrush}}{
            \includegraphics[width=0.45\textwidth]{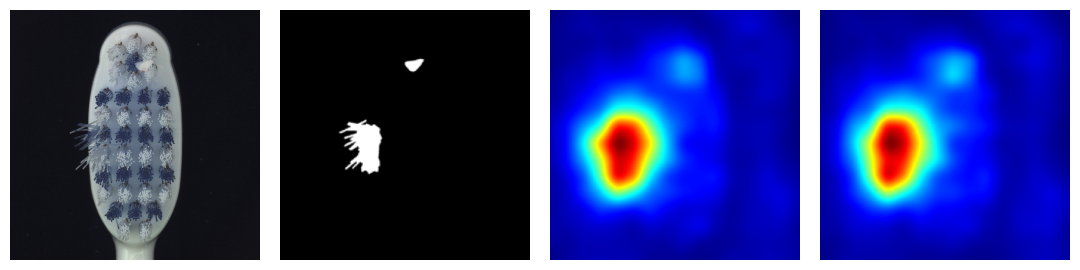}
        } \\
        \subcaptionbox{Transistor\label{fig:transistor}}{
            \includegraphics[width=0.45\textwidth]{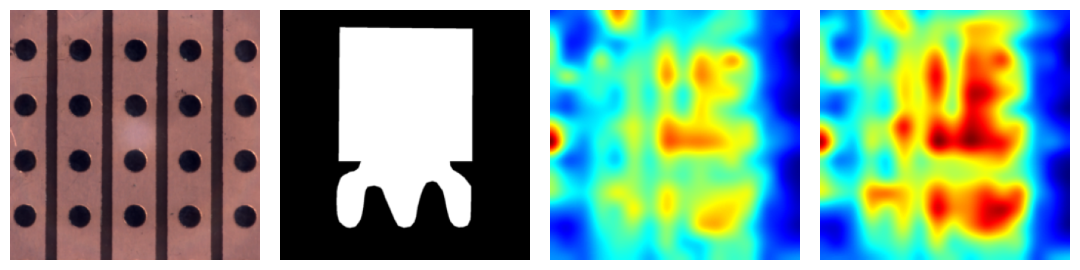}
        } &
        \subcaptionbox{Wood\label{fig:wood}}{
            \includegraphics[width=0.45\textwidth]{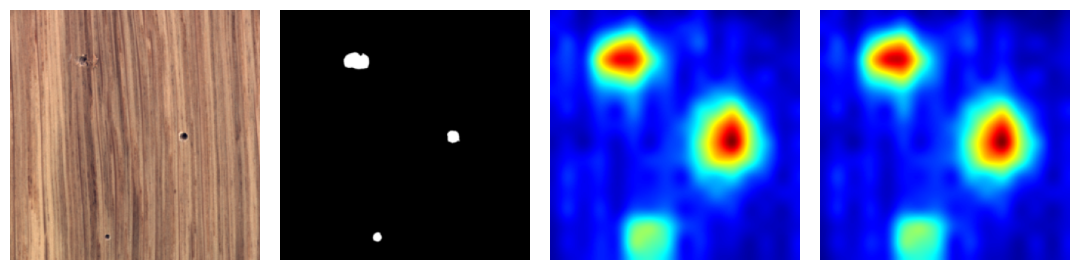}
        } \\
        \subcaptionbox{Zipper\label{fig:zipper}}{
            \includegraphics[width=0.45\textwidth]{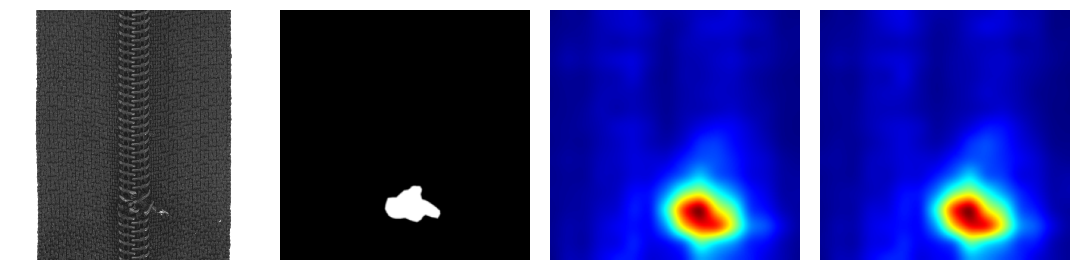}
        } &
        % 占位空格
        \\
    \end{tabular}
  \caption{Qualitative Comparison of Anomaly Detection Results Across 15 Categories Using the Baseline Method \textit{RD} and \textit{COAD}. Each subfigure shows: (from left to right) the original image, ground truth, result from the baseline method, and the result after applying \textit{COAD}.}
    \label{fig:rd_sm}
\end{figure*}

\begin{figure*}[htbp]
    \centering
    % 设置子图大小和行间距
    \setlength{\tabcolsep}{4pt} % 子图列间距
    \renewcommand{\arraystretch}{1.5} % 调整行间距

    \begin{tabular}{cc} % 两列布局
        \subcaptionbox{Bottle\label{fig:bottle}}{
            \includegraphics[width=0.45\textwidth]{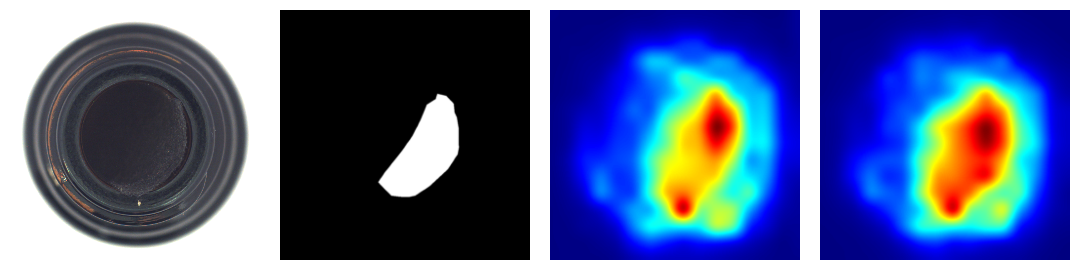}
        } &
        \subcaptionbox{Cable\label{fig:cable}}{
            \includegraphics[width=0.45\textwidth]{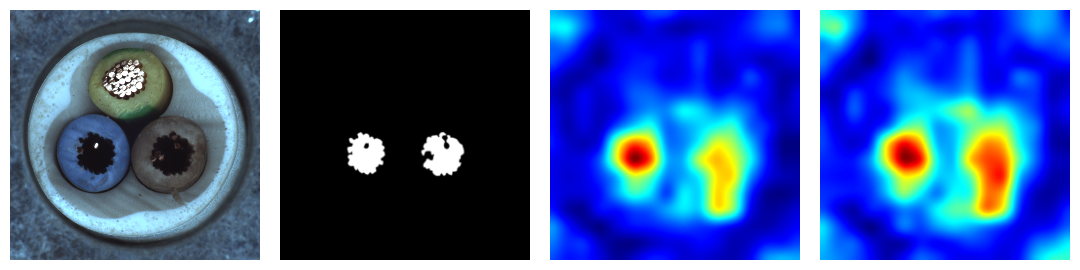}
        } \\
        \subcaptionbox{Capsule\label{fig:capsule}}{
            \includegraphics[width=0.45\textwidth]{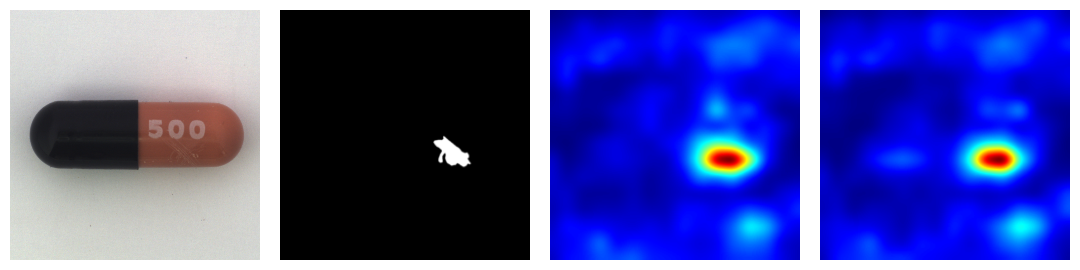}
        } &
        \subcaptionbox{Carpet\label{fig:carpet}}{
            \includegraphics[width=0.45\textwidth]{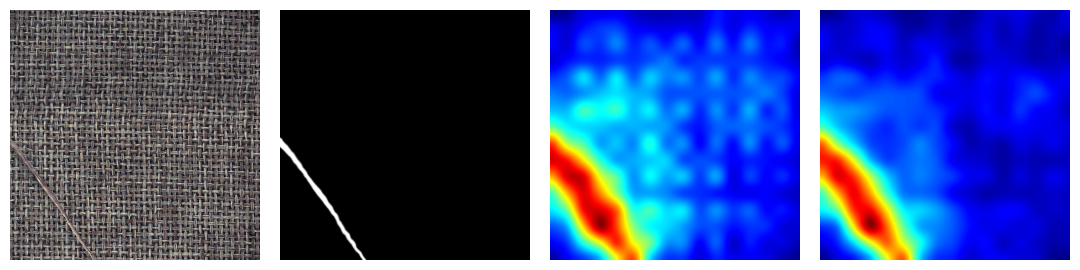}
        } \\
        \subcaptionbox{Grid\label{fig:grid}}{
            \includegraphics[width=0.45\textwidth]{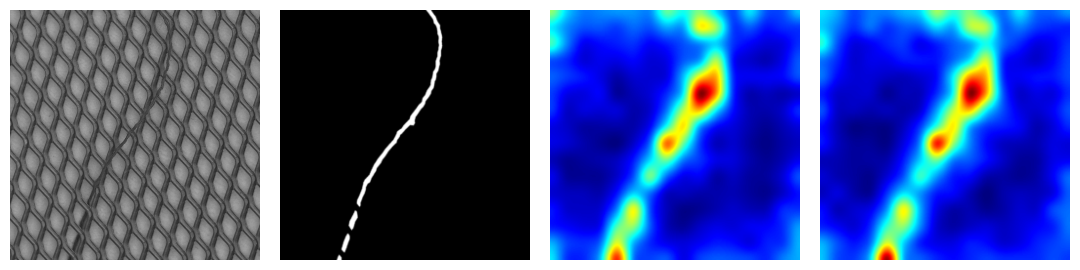}
        } &
        \subcaptionbox{Hazelnut\label{fig:hazelnut}}{
            \includegraphics[width=0.45\textwidth]{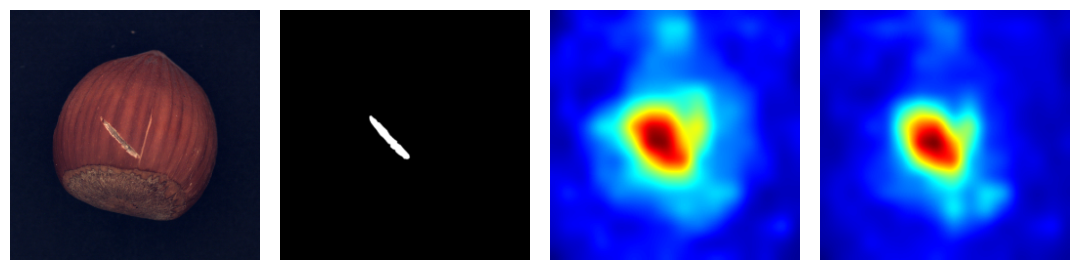}
        } \\
        \subcaptionbox{Leather\label{fig:leather}}{
            \includegraphics[width=0.45\textwidth]{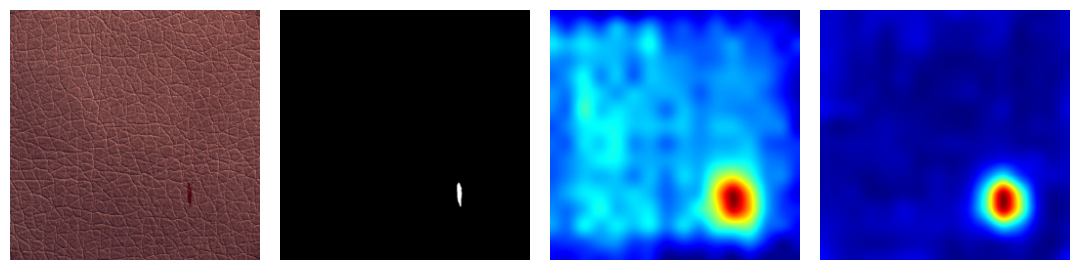}
        } &
        \subcaptionbox{Metalnut\label{fig:metal_nut}}{
            \includegraphics[width=0.45\textwidth]{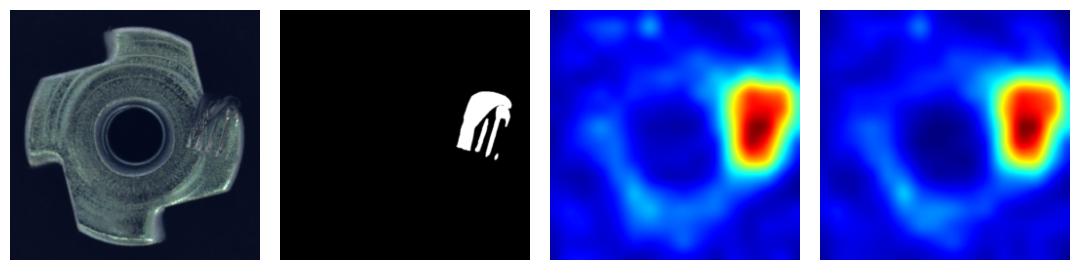}
        } \\
        \subcaptionbox{Pill\label{fig:pill}}{
            \includegraphics[width=0.45\textwidth]{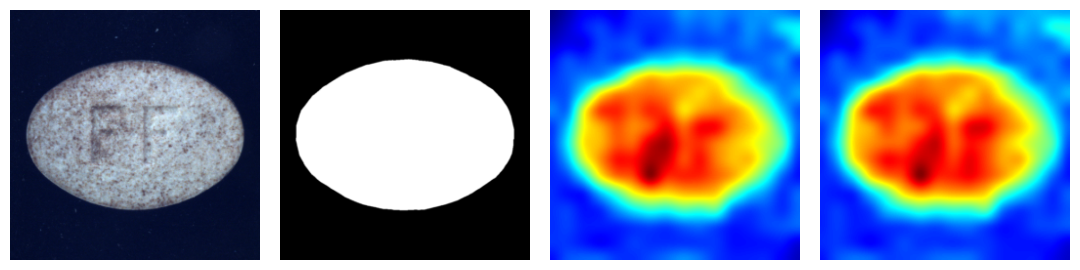}
        } &
        \subcaptionbox{Screw\label{fig:screw}}{
            \includegraphics[width=0.45\textwidth]{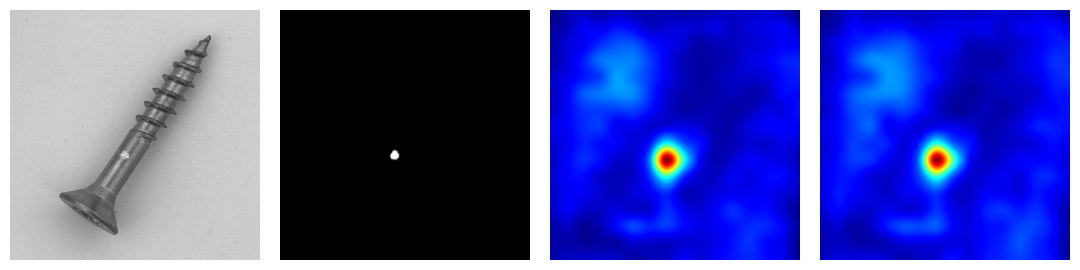}
        } \\
        \subcaptionbox{Tile\label{fig:tile}}{
            \includegraphics[width=0.45\textwidth]{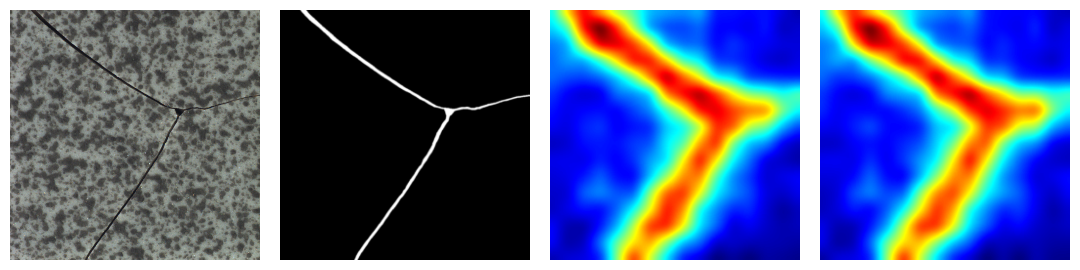}
        } &
        \subcaptionbox{Toothbrush\label{fig:toothbrush}}{
            \includegraphics[width=0.45\textwidth]{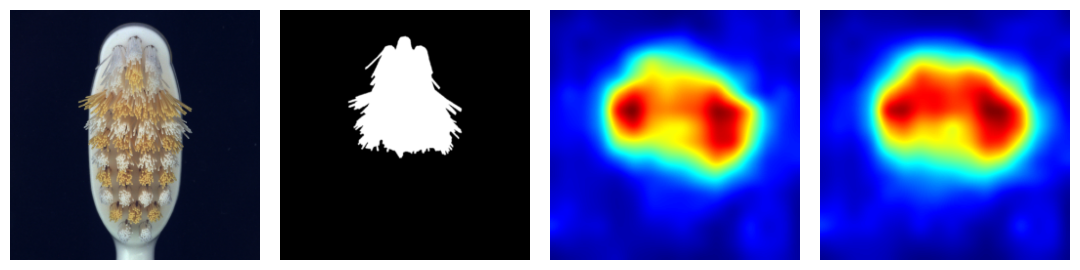}
        } \\
        \subcaptionbox{Transistor\label{fig:transistor}}{
            \includegraphics[width=0.45\textwidth]{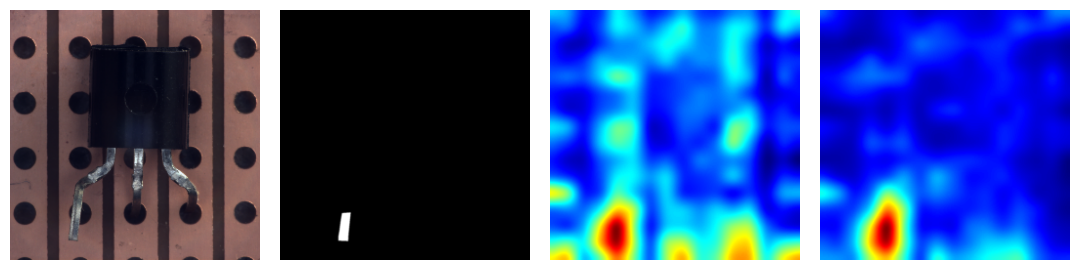}
        } &
        \subcaptionbox{Wood\label{fig:wood}}{
            \includegraphics[width=0.45\textwidth]{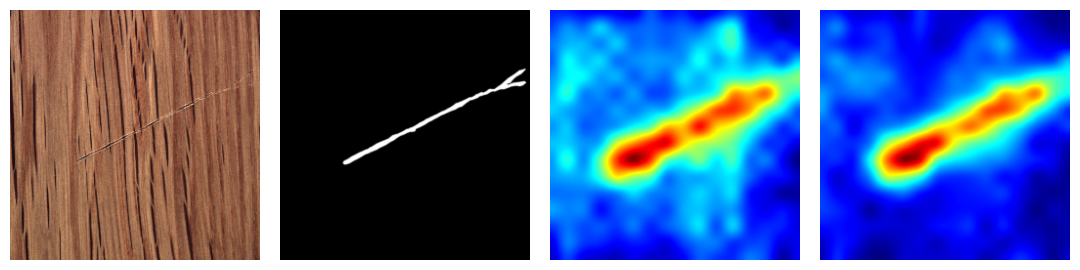}
        } \\
        \subcaptionbox{Zipper\label{fig:zipper}}{
            \includegraphics[width=0.45\textwidth]{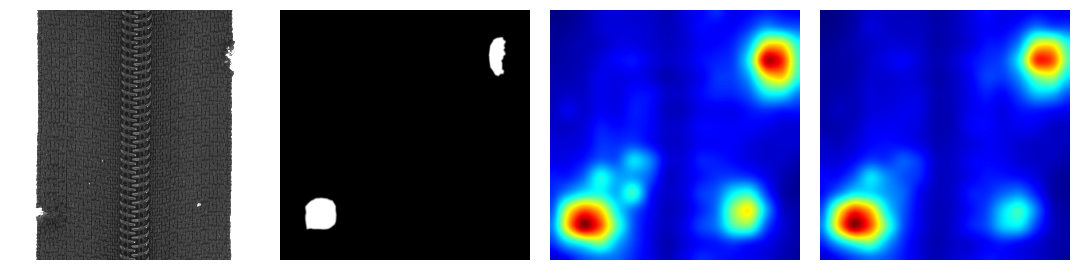}
        } &
        % 占位空格
        \\
    \end{tabular}
  \caption{Qualitative Comparison of Anomaly Detection Results Across 15 Categories Using a Baseline Method \textit{RD++} and \textit{COAD}. Each subfigure shows: (from left to right) the original image, ground truth, result from the baseline method, and the result after applying \textit{COAD}.}
    \label{fig:rd++_sm}
\end{figure*}

\begin{figure*}[htbp]
    \centering
    % 设置子图大小和行间距
    \setlength{\tabcolsep}{4pt} % 子图列间距
    \renewcommand{\arraystretch}{1.5} % 调整行间距

    \begin{tabular}{cc} % 两列布局
        \subcaptionbox{Bottle\label{fig:bottle}}{
            \includegraphics[width=0.45\textwidth]{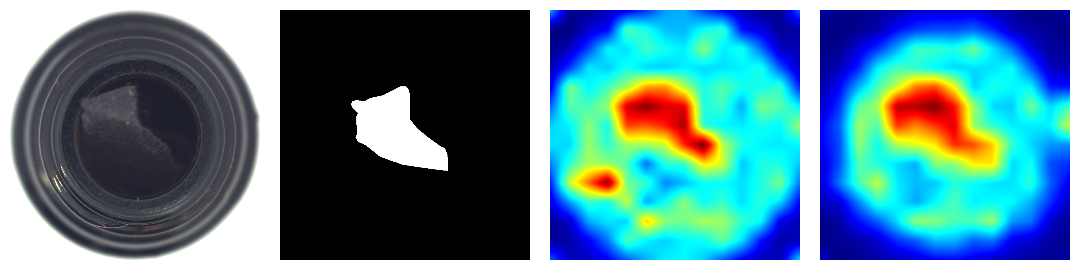}
        } &
        \subcaptionbox{Cable\label{fig:cable}}{
            \includegraphics[width=0.45\textwidth]{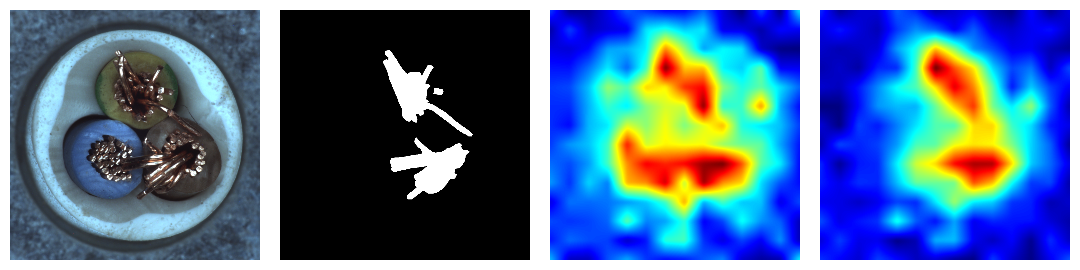}
        } \\
        \subcaptionbox{Capsule\label{fig:capsule}}{
            \includegraphics[width=0.45\textwidth]{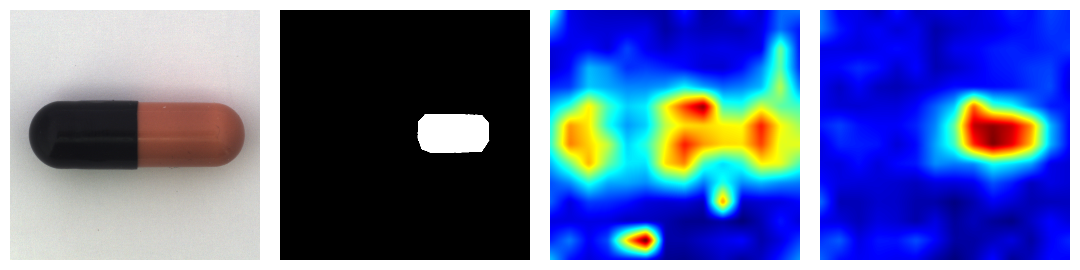}
        } &
        \subcaptionbox{Carpet\label{fig:carpet}}{
            \includegraphics[width=0.45\textwidth]{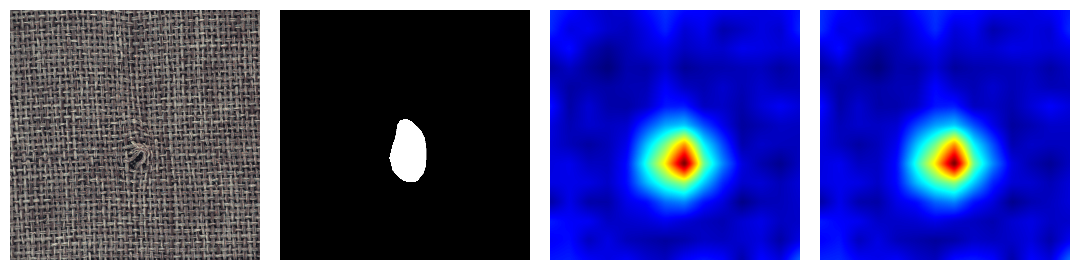}
        } \\
        \subcaptionbox{Grid\label{fig:grid}}{
            \includegraphics[width=0.45\textwidth]{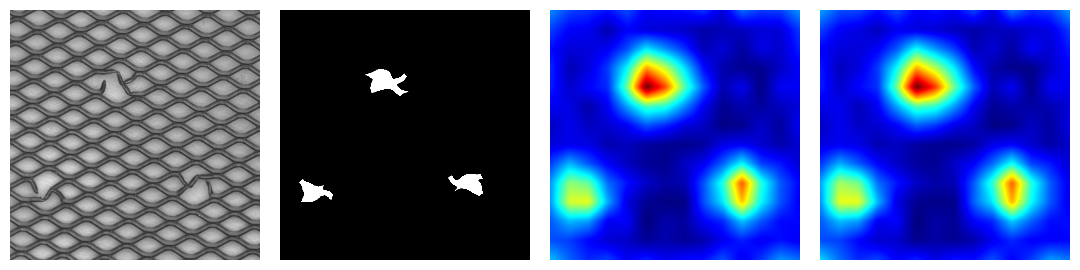}
        } &
        \subcaptionbox{Hazelnut\label{fig:hazelnut}}{
            \includegraphics[width=0.45\textwidth]{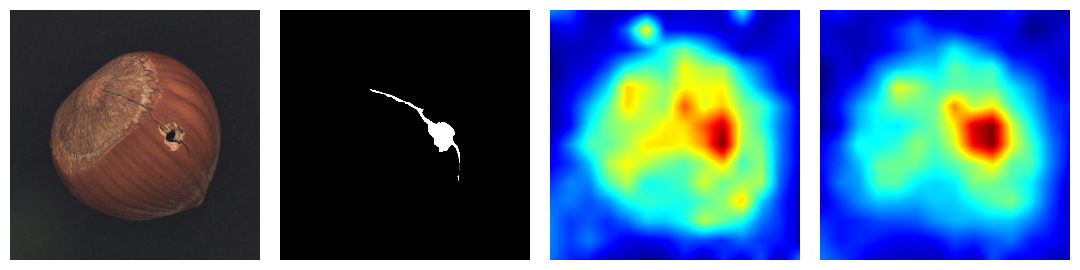}
        } \\
        \subcaptionbox{Leather\label{fig:leather}}{
            \includegraphics[width=0.45\textwidth]{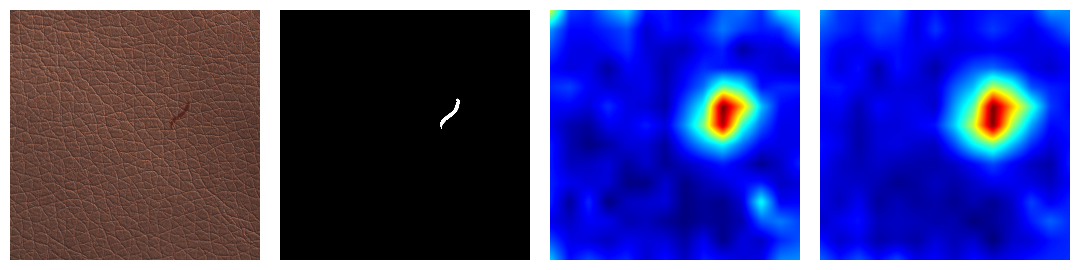}
        } &
        \subcaptionbox{Metalnut\label{fig:metal_nut}}{
            \includegraphics[width=0.45\textwidth]{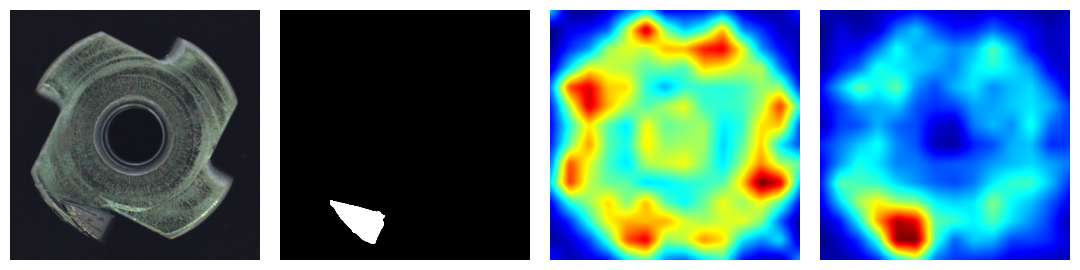}
        } \\
        \subcaptionbox{Pill\label{fig:pill}}{
            \includegraphics[width=0.45\textwidth]{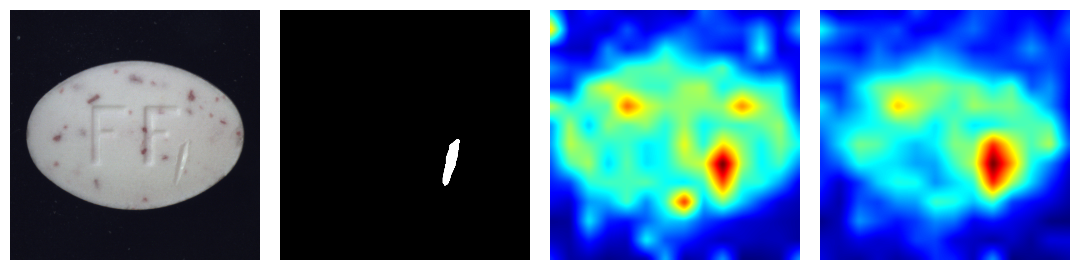}
        } &
        \subcaptionbox{Screw\label{fig:screw}}{
            \includegraphics[width=0.45\textwidth]{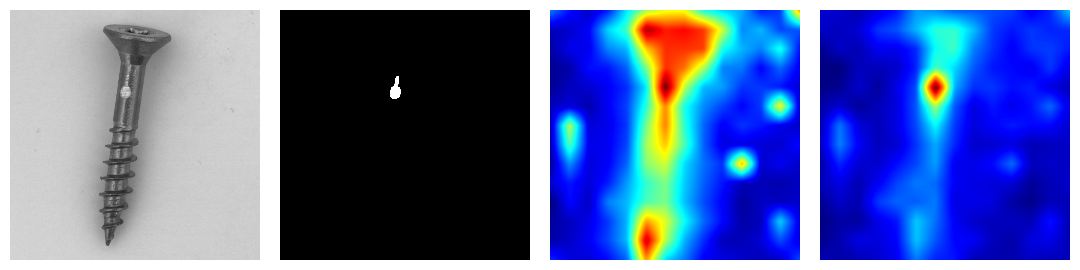}
        } \\
        \subcaptionbox{Tile\label{fig:tile}}{
            \includegraphics[width=0.45\textwidth]{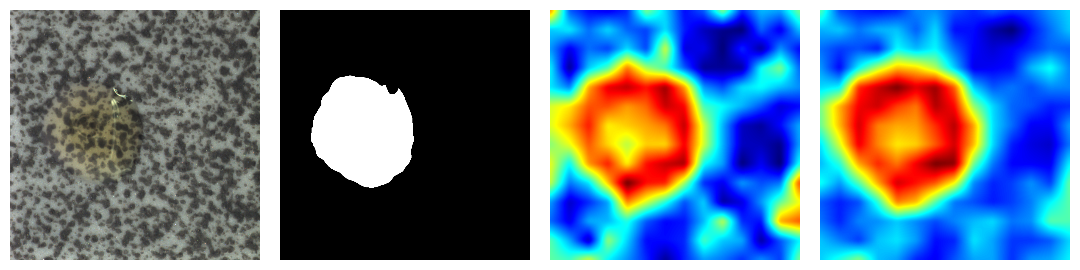}
        } &
        \subcaptionbox{Toothbrush\label{fig:toothbrush}}{
            \includegraphics[width=0.45\textwidth]{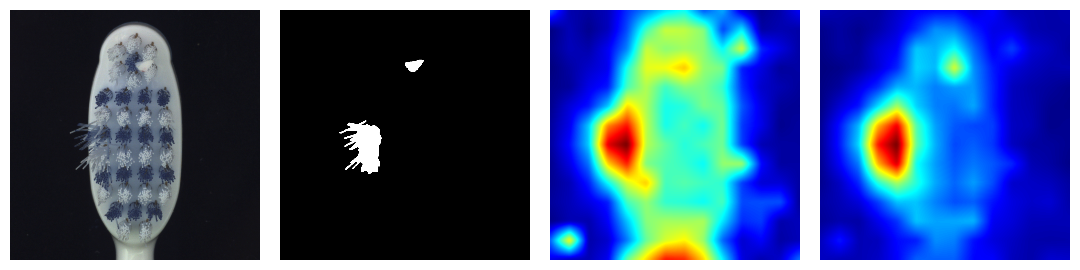}
        } \\
        \subcaptionbox{Transistor\label{fig:transistor}}{
            \includegraphics[width=0.45\textwidth]{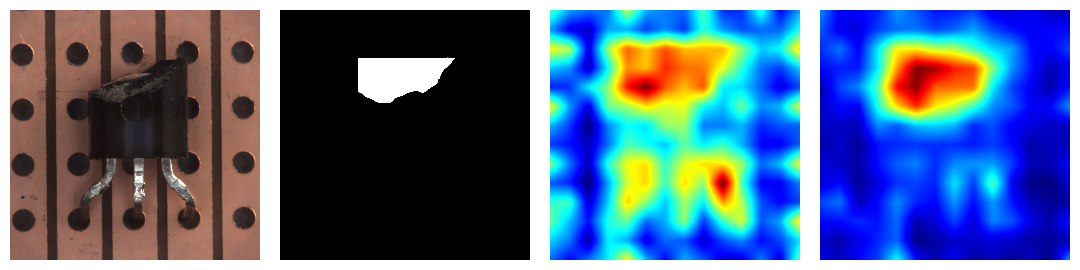}
        } &
        \subcaptionbox{Wood\label{fig:wood}}{
            \includegraphics[width=0.45\textwidth]{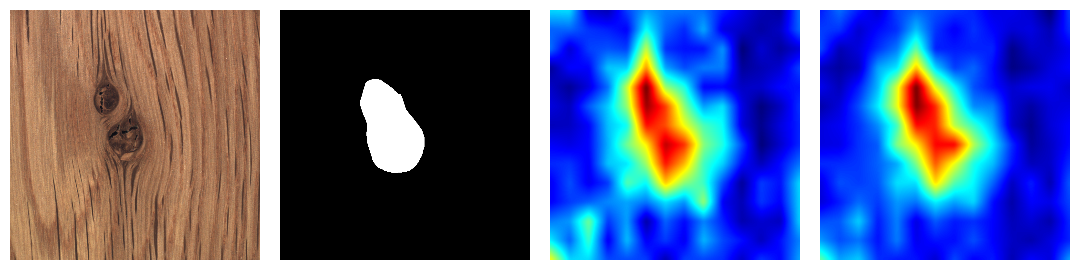}
        } \\
        \subcaptionbox{Zipper\label{fig:zipper}}{
            \includegraphics[width=0.45\textwidth]{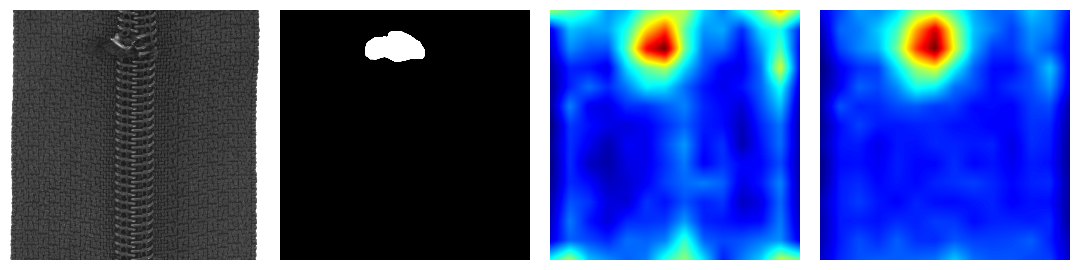}
        } &
        % 占位空格
        \\
    \end{tabular}
  \caption{Qualitative Comparison of Anomaly Detection Results Across 15 Categories Using a Baseline Method \textit{UniAD} and \textit{COAD}. Each subfigure shows: (from left to right) the original image, ground truth, result from the baseline method, and the result after applying \textit{COAD}.}
    \label{fig:uniad_sm}
\end{figure*}

\begin{figure*}[htbp]
    \centering
    % 设置子图大小和行间距
    \setlength{\tabcolsep}{4pt} % 子图列间距
    \renewcommand{\arraystretch}{1.5} % 调整行间距

    \begin{tabular}{cc} % 两列布局
        \subcaptionbox{Bottle\label{fig:bottle}}{
            \includegraphics[width=0.45\textwidth]{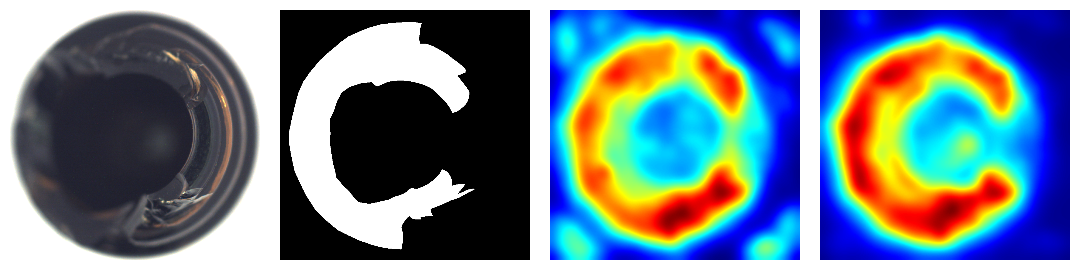}
        } &
        \subcaptionbox{Cable\label{fig:cable}}{
            \includegraphics[width=0.45\textwidth]{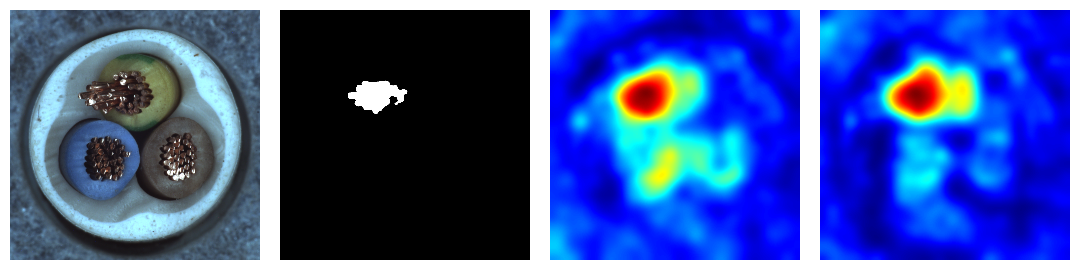}
        } \\
        \subcaptionbox{Capsule\label{fig:capsule}}{
            \includegraphics[width=0.45\textwidth]{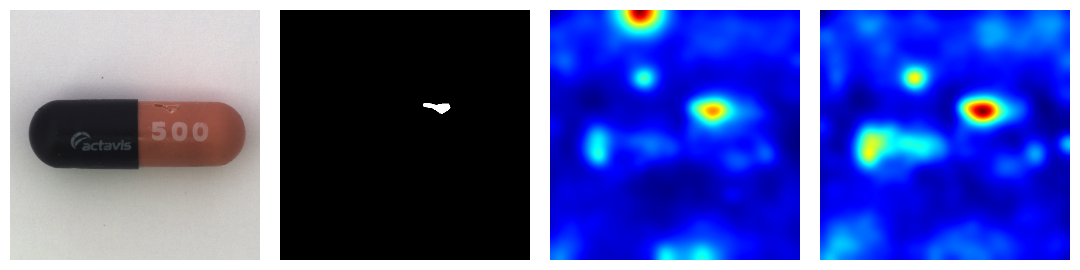}
        } &
        \subcaptionbox{Carpet\label{fig:carpet}}{
            \includegraphics[width=0.45\textwidth]{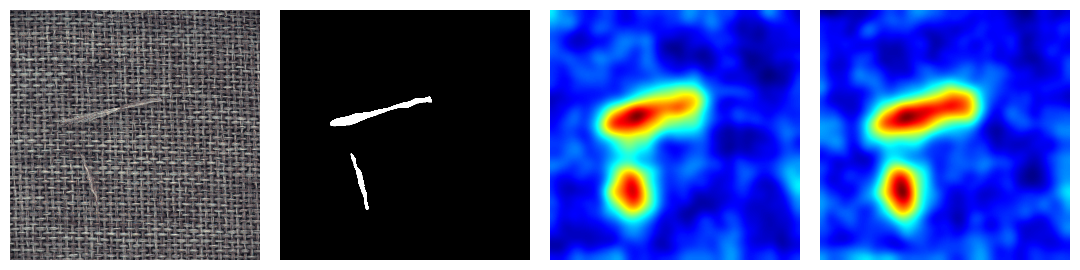}
        } \\
        \subcaptionbox{Grid\label{fig:grid}}{
            \includegraphics[width=0.45\textwidth]{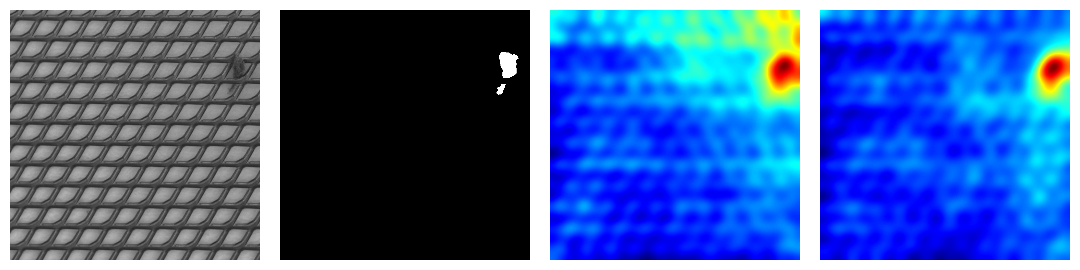}
        } &
        \subcaptionbox{Hazelnut\label{fig:hazelnut}}{
            \includegraphics[width=0.45\textwidth]{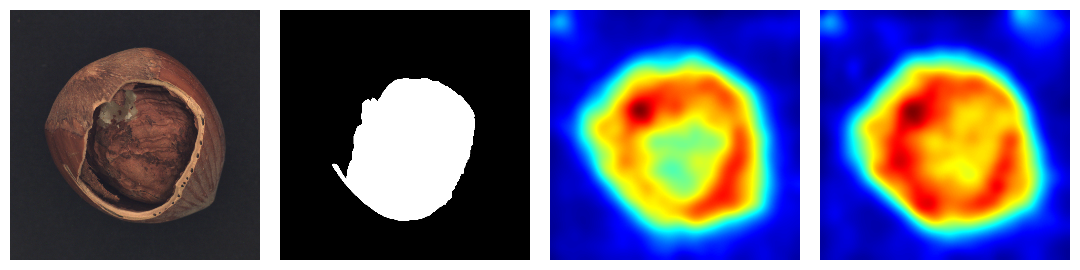}
        } \\
        \subcaptionbox{Leather\label{fig:leather}}{
            \includegraphics[width=0.45\textwidth]{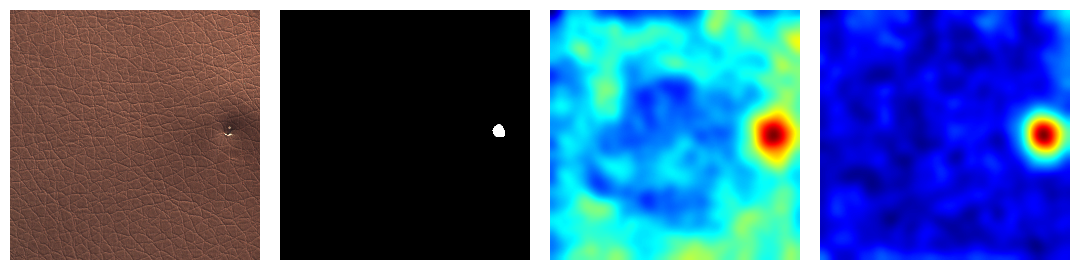}
        } &
        \subcaptionbox{Metalnut\label{fig:metal_nut}}{
            \includegraphics[width=0.45\textwidth]{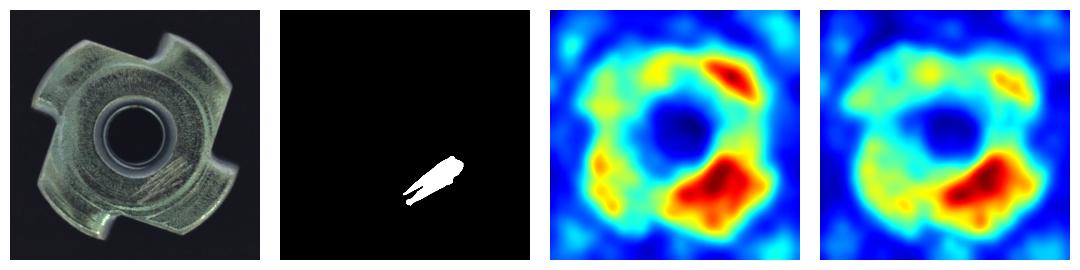}
        } \\
        \subcaptionbox{Pill\label{fig:pill}}{
            \includegraphics[width=0.45\textwidth]{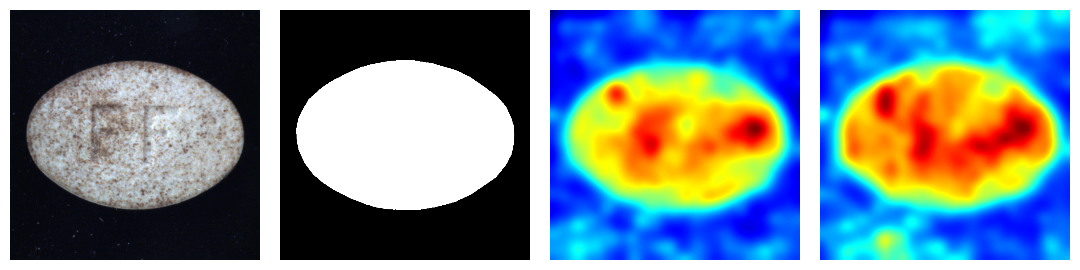}
        } &
        \subcaptionbox{Screw\label{fig:screw}}{
            \includegraphics[width=0.45\textwidth]{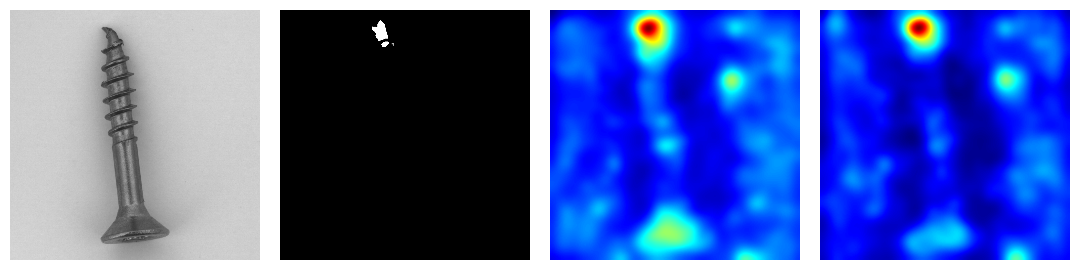}
        } \\
        \subcaptionbox{Tile\label{fig:tile}}{
            \includegraphics[width=0.45\textwidth]{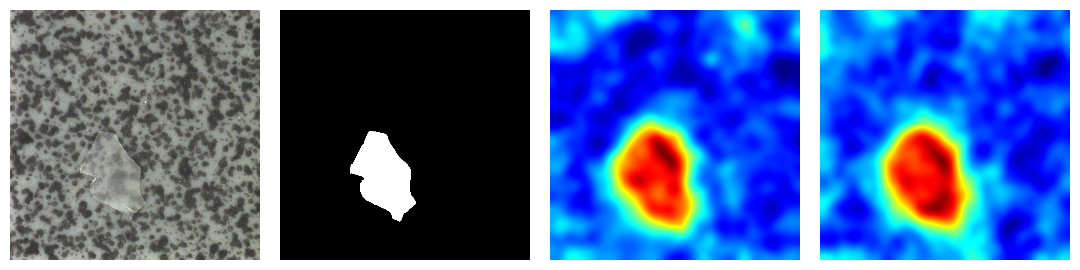}
        } &
        \subcaptionbox{Toothbrush\label{fig:toothbrush}}{
            \includegraphics[width=0.45\textwidth]{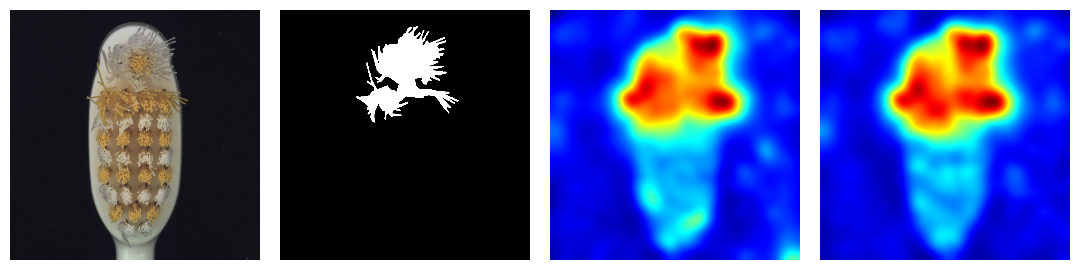}
        } \\
        \subcaptionbox{Transistor\label{fig:transistor}}{
            \includegraphics[width=0.45\textwidth]{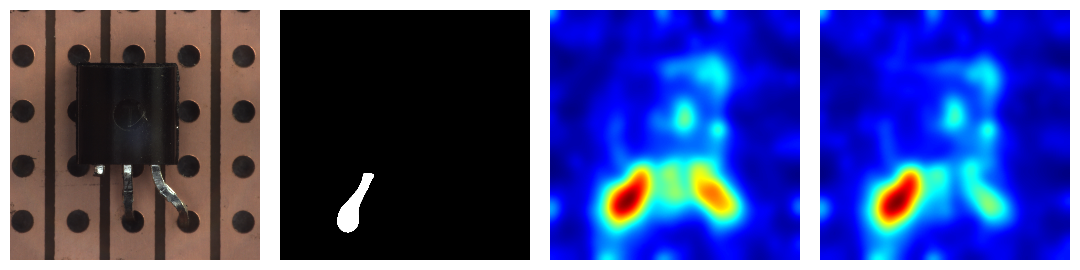}
        } &
        \subcaptionbox{Wood\label{fig:wood}}{
            \includegraphics[width=0.45\textwidth]{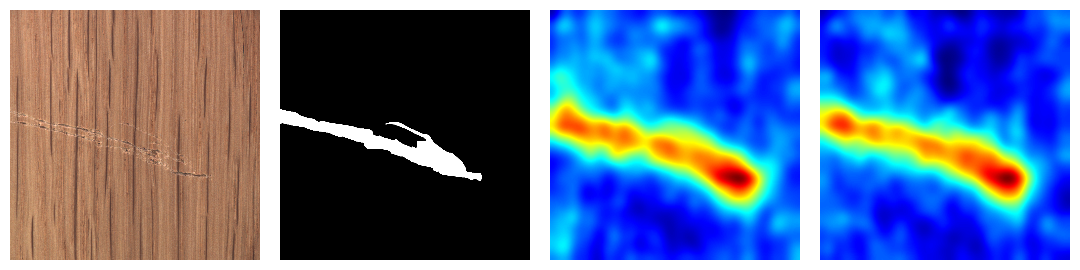}
        } \\
        \subcaptionbox{Zipper\label{fig:zipper}}{
            \includegraphics[width=0.45\textwidth]{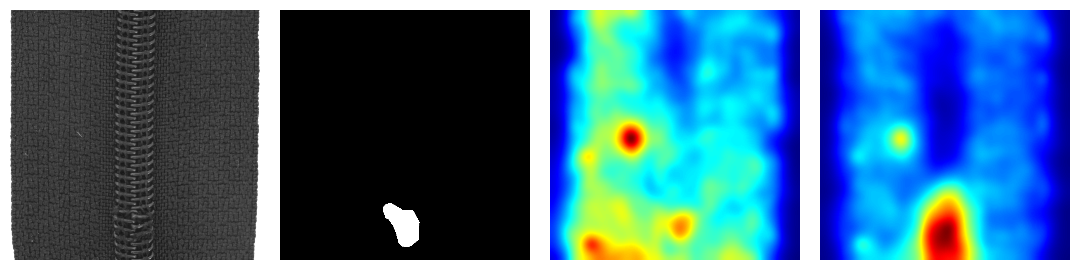}
        } &
        % 占位空格
        \\
    \end{tabular}
  \caption{Qualitative Comparison of Anomaly Detection Results Across 15 Categories Using a Baseline Method \textit{DiAD} and \textit{COAD}. Each subfigure shows: (from left to right) the original image, ground truth, result from the baseline method, and the result after applying \textit{COAD}.}
    \label{fig:diad_sm}
\end{figure*}

To further validate the effectiveness of our methods, we provide a qualitative comparison of anomaly detection results across 15 distinct categories using baselines such as \textit{RD}, \textit{RD++}, \textit{UniAD}, \textit{DiAD}, and the proposed \textit{COAD} framework in ~\cref{fig:rd_sm,fig:rd++_sm,fig:uniad_sm,fig:diad_sm}. Each subfigure demonstrates: (from left to right) the original image, the ground truth anomaly mask, the result from the baseline method, and the result after applying \textit{COAD}.

The comparison shows that \textit{COAD} effectively reduces both false positive rates (FPR) and false negative rates (FNR), accurately capturing anomaly regions while minimizing misclassifications. These results are consistent with the improvements reflected in the earlier quantitative data in ~\cref{tab:auroc_image,tab:auroc_pixel}, where \textit{COAD} consistently outperforms its counterparts in image-level and pixel-level AUROC metrics, highlighting \textit{COAD}'s capability to outperform the baselines. 

By leveraging controllable overfitting, \textit{COAD} enhances the model's sensitivity to true anomalies while maintaining its generalization capabilities. The reduction in FPR minimizes the misclassification of normal regions as anomalies, while the decreased FNR highlights the model's ability to capture subtle anomalies that baseline methods often miss. These qualitative and quantitative improvements collectively demonstrate the robustness and effectiveness of \textit{COAD} in advancing anomaly detection tasks. Additionally, the sharper distinction between normal and anomalous regions demonstrates the framework’s capacity to refine boundary precision, further enhancing detection reliability.

\section{Future Work}

In future work, we plan to refine our pseudo-anomaly generators by incorporating advanced methods in ~\cref{sec:glass-like}, enabling the generation of more structured and domain-relevant pseudo-anomalies.

Moreover, we intend to explore the application of the \textit{COAD} framework in other domains, particularly in areas such as medical imaging, autonomous driving, and forgery detection. Forgery detection, for instance, highlights an opportunity to utilize \textit{COAD}'s enhanced sensitivity to subtle anomalies, such as fine-grained pattern inconsistencies or texture deviations. Furthermore, we aim to generalize and modularize the \textit{COAD} framework into a plug-and-play component that can be seamlessly integrated into similar tasks across various fields. Ultimately, we aspire to redefine the concept of overfitting itself, transforming it from a perceived limitation into a powerful tool that drives innovation.

\end{document}